\documentclass{article}

\usepackage[final]{neurips_2022}

\usepackage[utf8]{inputenc} 
\usepackage[T1]{fontenc}    
\usepackage{hyperref}       
\usepackage{url}            
\usepackage{booktabs}       
\usepackage{amsfonts}       
\usepackage{nicefrac}       
\usepackage{microtype}      
\usepackage{xcolor}         
\usepackage{amsmath}
\usepackage{caption}
\usepackage{subcaption}
\usepackage{graphicx}
\usepackage{float}
\usepackage{amsthm}

\newcommand{\E}{\mathbb{E}}
\newcommand{\kl}{\text{KL}}

\newcommand{\elbo}{\text{ELBO}}
\newcommand{\dlm}{\text{DLM}}

\newcommand{\erm}{\text{ERM}}
\newcommand{\set}[1]{\mathcal{#1}}
\newcommand{\bigO}{\mathcal{O}}
\newcommand{\myappendix}{Appendix}

\newtheorem{theorem}{Theorem}
\newtheorem{proposition}[theorem]{Proposition}

\newtheorem{observation}{Observation}

\DeclareMathOperator*{\argmin}{arg\,min}

\usepackage{natbib}
\PassOptionsToPackage{options}{natbib}

\title{On the Performance of Direct Loss Minimization for Bayesian Neural Networks}

\author{ 
{\bf Yadi Wei} \\ Indiana University \\ Bloomington, IN \\ \texttt{weiyadi@iu.edu}
\And
{\bf Roni Khardon} \\ Indiana University \\ Bloomington, IN \\ \texttt{rkhardon@iu.edu}
}

\begin{document}

\maketitle

\begin{abstract}
  Direct Loss Minimization (DLM) has been proposed as a pseudo-Bayesian method motivated as regularized loss minimization. Compared to variational inference, it replaces the loss term in the evidence lower bound (ELBO) with the predictive log loss, which is the same loss function used in evaluation. 
  A number of theoretical and empirical results in prior work suggest that DLM can significantly improve over ELBO optimization for some models. However, 
  as we point out in this paper, this is not the case for 
  Bayesian neural networks (BNNs).
  The paper explores the practical performance of DLM for BNN, 
  the reasons for its failure and its relationship to optimizing the ELBO,
  uncovering some interesting facts about both algorithms.
\end{abstract}

\section{Introduction}
One of the main goals of probabilistic machine learning is to develop algorithms that can make well calibrated probabilistic predictions. 
From the frequentist view, we need to find a \emph{single} set of parameters that best fits the data; while from a Bayesian view, we specify a prior distribution on the parameters, calculate the posterior, and then use the posterior to predict on new data. Let $D=\{(x^{(i)}, y^{(i)})\}_{i=1}^N$ be the dataset sampled i.i.d. from distribution $\set D$. From now on, we use superscript with parentheses to denote the $i$-th instance. Let $\theta$ denote the parameters. The frequentist method chooses one best set of parameters $\theta^*$ and makes predictions on new data $x^*$ such that $p(y^*|x^*)=p(y^*|\theta^*, x^*)$.  
Bayesian methods specify a prior $p(\theta)$, calculate the posterior
\begin{align*}
    p(\theta|D) \propto p(\theta) p(D|\theta),
\end{align*}
and make predictions on new data $x^*$ as
\begin{align*}
    p(y^*|x^*)=\E_{p(\theta|D)} [p(y^*|\theta, x^*)].
\end{align*}
For simple models, the posterior can be computed analytically. But for complicated models, the posterior becomes intractable. One solution is to get samples from true posterior and calculate the objective, for example using MCMC. Another line of work aims to find a distribution $q$ from an analytical distribution family $\set Q$ that is closest to the posterior. When making prediction on new data $x^*$, we marginalize over $q$, that is, $p(y^*|x^*)=\E_{q(\theta)}[p(y^*|\theta, x^*)]$.
A typical example is variational inference, which tries to minimize the KL divergence between the variational distribution and the true posterior:
\begin{align}
\label{eq:elbo}
    q^*(\theta) &= \argmin_{q\in \set Q}\kl(q(\theta) || p(\theta | D)) \nonumber \\ 
    &= \argmin_{q\in \set Q} \E_{q(\theta)}[\log q(\theta) - \log p(\theta, D)] + \log p(D) \nonumber \\
    &=\argmin_{q\in \set Q} \sum_i
    \E_{q(\theta)}[-\log p(y^{(i)}|\theta, x^{(i)})] + \kl(q(\theta) || p(\theta)).
\end{align}
The last line is the negation of the 
Evidence Lower Bound (ELBO).
The most common measure to evaluate the quality of predictions is negative log-loss (NLL):
\begin{align}
\label{eq:log-loss}
    l(q, (x^*, y^*)) = -\log \E_{q(\theta)}[p(y^*|\theta, x^*)].
\end{align}
We use $l_{\text{test}}(q)$ to denote the averaged NLL on test set.
We optimize eq \eqref{eq:elbo} but hope to get lower NLL, i.e. eq \eqref{eq:log-loss}. 
This suggests a discrepancy.
If we care about the NLL, why not directly optimize NLL (\ref{eq:log-loss})? From this perspective the KL term in eq \eqref{eq:elbo} is seen as a regularizer to prevent overfitting. This motivates the idea of Direct Loss Minimization (DLM) which has been studied by multiple authors:
\begin{align}
\label{eq:dlm}
    q_{\dlm}^{(\eta)}(\theta) = \argmin_{q\in \set Q} \sum_i 
    -\log \E_{q(\theta)}[p(y^{(i)}|\theta, x^{(i)})] + \eta\kl(q(\theta)||p(\theta)).
\end{align}
Notice that by Jensen's inequality,
\begin{align*}
    l_{\text{dlm}}(q, (x, y)) = -\log \E_{q(\theta)}[p(y|\theta, x)] \leq l_{\text{elbo}}(q, (x, y)) = -\E_{q(\theta)} [\log p(y|\theta, x)]
\end{align*}
so that $l_{\text{elbo}}$ can be seen as a surrogate loss for the log loss which is used during training. But DLM optimizes the desired objective and 
the idea of DLM can be applied on various loss functions. We can replace the first term in eq \eqref{eq:dlm} with any loss functions we are interested in. 
Similarly, the regularizer can be chosen among several options. However, in this paper we focus on the choice given above.


In practice, it is common to use a hyperparameter $\eta$ to tune the KL regularizer, as in eq \eqref{eq:dlm}.
In addition, when it is hard to compute the expectations analytically we can use Monte Carlo samples to approximate them.
With these modifications
the objectives in eq \eqref{eq:elbo} and eq \eqref{eq:dlm}) become:
\begin{align}
    \label{eq:eta-elbo-mc}
    \bar{q}_{\elbo}^{(\eta, M)}(\theta) &= \argmin_{q\in \set Q} \sum_i
    \frac{1}{M} \sum_{m=1}^M [-\log p(y^{(i)}|\theta^{(m)}, x^{(i)})] + \eta \kl(q(\theta) || p(\theta)), &\theta^{(m)}\sim q(\theta); \\
    \label{eq:eta-dlm-mc}
    \bar{q}_{\dlm}^{(\eta, M)}(\theta) &= \argmin_{q\in \set Q} \sum_i 
    -\log \frac{1}{M} \sum_{m=1}^M p(y^{(i)}|\theta^{(m)}, x^{(i)}) + \eta \kl(q(\theta)||p(\theta)), &\theta^{(m)}\sim q(\theta).
\end{align}
Notice that eq \eqref{eq:eta-elbo-mc} is an unbiased estimate of eq \eqref{eq:elbo} while eq \eqref{eq:eta-dlm-mc} is a \emph{biased} estimate of eq \eqref{eq:dlm}.

\section{Theoretical Motivation of DLM}
\label{sec:theory}

Prior work motivates the use of DLM from a theoretical perspective. In this section we review some of these results.  
Specifically, 
\cite{Sheth2019pseudo} provide risk bounds for several variants of DLM. Here we focus on one result
(see their appendix B) that uses a bounded optimization view. Suppose we restrict our distribution family $\set Q$ to $\set Q_A=\{q\in \set Q \text{ s.t. } \kl(q, p)\leq A\}$, where $p$ is the prior distribution over $\theta$, and we perform empirical risk minimization (ERM) on $\set Q_A$:
\begin{align}
    \label{eq:erm}
    q_{\erm}^{(A)}(\theta) = \argmin_{q\in \set Q_A} 
    \sum_i
    -\log \E_{q(\theta)}[p(y^{(i)}|\theta, x^{(i)})].
\end{align}
Then
with probability $1-\delta$ over the choice of the dataset $D$, for all $q\in \set Q_A$, 
{\small
\begin{align}
\label{ineq:erm}
    \E_{(x,y)\sim \set D}\left[-\log \left(\E_{q_\erm^{(A)}(\theta)} p(y|\theta, x) \right) \right] \leq \E_{(x,y)\sim \set D} \left[-\log (\E_{q(\theta)}(p(y|\theta, x))) \right] + \bigO\left(\sqrt{\frac{A}{N}} + \sqrt{\frac{\log \frac{1}{\delta}}{N} }\right).
\end{align}
}
This result holds when the log-loss is bounded, which
can be achieved by replacing the log loss with 
\begin{align}
\label{eq:smooth}
    \log^{(a)} p = \log((1-a)p + a)    
\end{align}
or by further bounding the parameter space of distribution family $\set Q$.
The following proposition (proof in \myappendix) observes
that while the above holds for eq \eqref{eq:erm} we can obtain similar risk bounds for eq \eqref{eq:dlm}.
One can further extend eq \eqref{ineq:erm} into a data dependent bound as in Theorem 10 in \citep{Meir-Zhang}.
Therefore DLM as motivated above enjoys some theoretical support.
\begin{proposition}
Let $A_\eta=\kl(q_{\dlm}^{(\eta)} || p)$.
Then the solution of eq \eqref{eq:dlm}, i.e., $q_{\dlm}^{(\eta)}$, is also the solution of eq \eqref{eq:erm} with $A=A_\eta$. 
\end{proposition}

A second theoretical perspective is given by the 
recent work of \cite{pac-m}.
This work presents a PAC-Bayes bound 
called $\text{PAC}^m$ bound. With probability $1-\delta$, for any $q\in \set Q$,
\begin{align}
\label{ineq:pac-m}
    & \E_{(x, y) \sim \set D} [-\log \E_{q(\theta)}[p(y|\theta, x)]] \nonumber \\
    \leq& -\frac{1}{N}\sum_i \E_{q(\theta^M)} \left[\log \left(\frac{1}{M}\sum_m p\left(y^{(i)}|x^{(i)}, \theta^{(m)}\right)\right) \right] + \frac{1}{\eta N} \kl(q||p) \nonumber \\
    &+\psi(\set D, \eta, M, N, p, \delta) + \frac{1}{\eta M N} \log \frac{1}{\delta},
\end{align}
where 
{\small
\begin{align*}
    &\psi(\set D, \eta, M, N, p, \delta) = \frac{1}{\eta M N} \log \E_{D \sim \set D^N} \E_{p(\theta^{(1:m)})}\left [\exp\left(\eta N M \cdot \Delta\left(D, \theta^{(1:M)}\right)\right)\right], \\
    &\Delta(D, \theta^{(1:M)}) = \frac{1}{N} \sum_i \log \left(\frac{1}{M} \sum_m p(y^{(i)}|x^{(i)}, \theta^{(m)}) \right) - \E_{(x, y) \sim \set D} \left[\log \left( \frac{1}{M} 
    \sum_m p(y|x, 
    \theta^{(m)}) \right) \right].
\end{align*}
}
The $\text{PAC}^m$ algorithm minimizes the right-hand-side of \eqref{ineq:pac-m} to calculate its solution which we denote as $\bar{q}_{\dlm}^{(\eta)}$.
For the corresponding algorithm note that the terms on the last line of \eqref{ineq:pac-m} do not depend on $\bar{q}_{\dlm}^{(\eta)}$ and can be omitted in the optimization.
This is different from previous analysis \citep{sheth2017excess, Sheth2019pseudo}, that uses the predictive loss
\begin{align}
\label{eq:predictive-loss}
    -\log \E_{q(\theta)} [p(y|x, \theta)]&=-\log \E_{q(\theta^{(1:M)})}\left[\frac{1}{M} \sum_m p(y|x, \theta^{(m)})\right] \nonumber \\
    &\leq -\E_{q(\theta^{(1:M)})} \left[\log \left(\frac{1}{M} \sum_m p\left(y|x, \theta^{(m)}\right) \right) \right]
\end{align}
for the data-dependent upper bound. 
When the outside expectation in \eqref{ineq:pac-m} is implemented with a single multi-sample from $q()$,
which is the case in \citep{pac-m},
eq \eqref{ineq:pac-m} leads to the same implementation as eq \eqref{eq:eta-dlm-mc} and \eqref{ineq:erm}. 
Theoretically, loss term in bound \eqref{ineq:erm} is lower but it requires $\bar{q}_{\dlm}^{(\eta)}$ to converge to $q_{\dlm}^{(\eta)}$ to guarantee the performance (see related analysis by \citet{dlm-sgp}), while the bound \eqref{ineq:pac-m} directly guarantees the performance of $\bar{q}_{\dlm}^{(\eta)}$ with a higher loss as shown in eq \eqref{eq:predictive-loss}.

\cite{pac-m} also establishes the relationship between ELBO and DLM. Notice that with $M=1$, the right hand sides of eq \eqref{eq:eta-elbo-mc} and eq \eqref{eq:eta-dlm-mc} are the same. They also show that as $M$ becomes larger, the data dependent bound when $M>1$ (i.e., the right hand side of eq \eqref{ineq:pac-m}) is tighter than that when $M=1$, corresponding to the data dependent bound for ELBO. 

\section{Applications of DLM}

DLM has already been applied in practice and is shown to yield good results. \cite{sheth2017excess} apply DLM to the correlated topic model and it achieves lower predictive loss than ELBO. \cite{Jankowiak2020ParametricGP} explores the application of DLM in conjugate sparse Gaussian processes and \cite{dlm-sgp} extends this to non-conjugate sparse Gaussian processes. In those experiments, $\eta$ is chosen appropriately through cross validation. 
For conjugate cases, where both the ELBO objective \eqref{eq:elbo} and the DLM objective \eqref{eq:dlm} can be computed exactly without approximation, DLM significantly outperforms ELBO; while for non-conjugate cases, Monte Carlo sampling is needed and DLM is better than or comparable to ELBO.

DLM has also been applied to BNNs. \cite{bnn-rank1} apply both ELBO and DLM on a rank-1 parametrization of BNNs that they introduce. Their experiments show that DLM has higher NLL than ELBO, which means that DLM performs worse than ELBO. \cite{pac-m} also compare ELBO and DLM on BNNs and conclude that DLM performs better than ELBO when data is misspecified, i.e. the data generating distribution is not inside the model space. To see this, they apply ELBO and DLM to independently predict pixels of the bottom half of an image given the top half, which encounters data misspecification as the pixels are not independent. When there is no evident data misspecification they show that, under the same KL value, DLM performs slightly better than ELBO. However, the converged ELBO solution may not have the same KL value as the converged DLM solution. So the experiment is not sufficient to show that the converged DLM solution performs better than the converged ELBO solution.

\section{DLM in Bayesian Neural Networks}
\label{dlm-bnn}
\begin{figure}[H]
    \centering
    \includegraphics[width=0.7\textwidth]{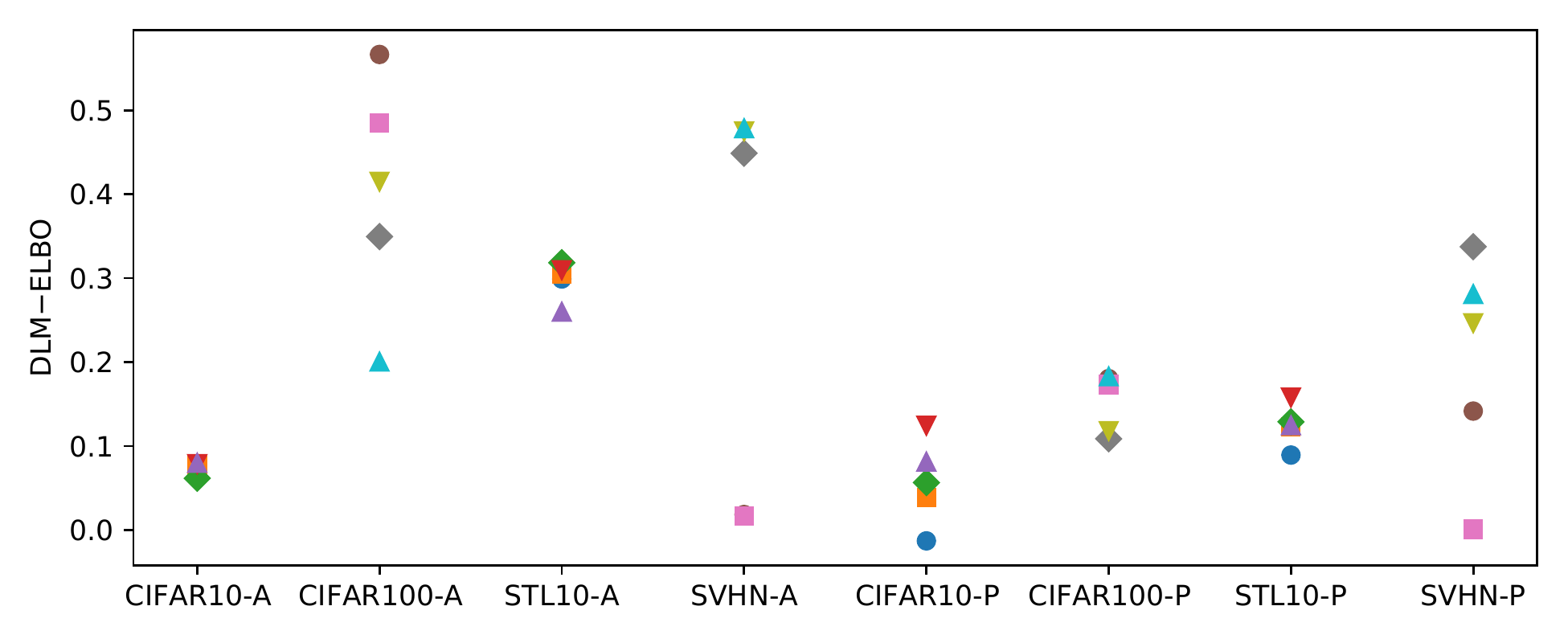}
    \caption{Comparison of ELBO and DLM in all experiments, ``-A'' means on AlexNet and ``-P'' means on PreResNet20. $y$-axis is $l_{\text{test}}(q_\dlm) - l_{\text{test}}(q_\elbo)$, and
    a positive value indicates that DLM performs worse than ELBO. Each point represents an independent run with random initialization.}
    \label{fig:all-result}
\end{figure}

In contrast with the positive evidence, our work found that DLM does not perform as well for Bayesian neural networks (BNNs).
In BNN,
$\theta$ represents the weights of the neural network and we use eq \eqref{eq:eta-elbo-mc} and eq \eqref{eq:eta-dlm-mc} with a mean-field diagonal Gaussian variational distribution $q(\theta_j)=\mathcal{N}(\mu_j, \sigma_j^2)$ and $\mu$ and $\sigma^2$ are vectors of the same dimension as the parameter space. BNNs normally have millions of parameters, so the KL-divergence term can be very high and using a high value of $\eta=1$ leads to poor performance,
hence we fix $\eta=0.1$. Following \cite{competition-approximate-inference}, we set the prior variance to 0.05. 
We experimented with two neural network structures, AlexNet \citep{alexnet} and PreResNet \citep{preresnet} with depth 20 on four datasets, CIFAR10, CIFAR100, STL10, SVHN. 
For all experiments, we set the batch size to 512, use the Adam optimizer with learning rate 0.001 and train for 500 epochs. We set $M=5$ in eq \eqref{eq:eta-elbo-mc} and eq \eqref{eq:eta-dlm-mc} for training and $M=10$ for evaluation.
Most combinations of datasets and structures have similar results (see Figure \ref{fig:all-result}) and behave similarly during our exploration, so we only show the detailed exploration of using AlexNet on CIFAR10 in the main body of the paper and discuss exceptions in the \myappendix.
To simplify the presentation, we use upper case ``ELBO'' and ``DLM'' to indicate the solution trained with ELBO and DLM loss respectively, and use lower case ``elbo'' and ``dlm'' to denote the corresponding loss functions.

\begin{figure}[H]
    \centering
    \begin{subfigure}[t]{0.32\textwidth}
         \centering
         \includegraphics[width=\textwidth]{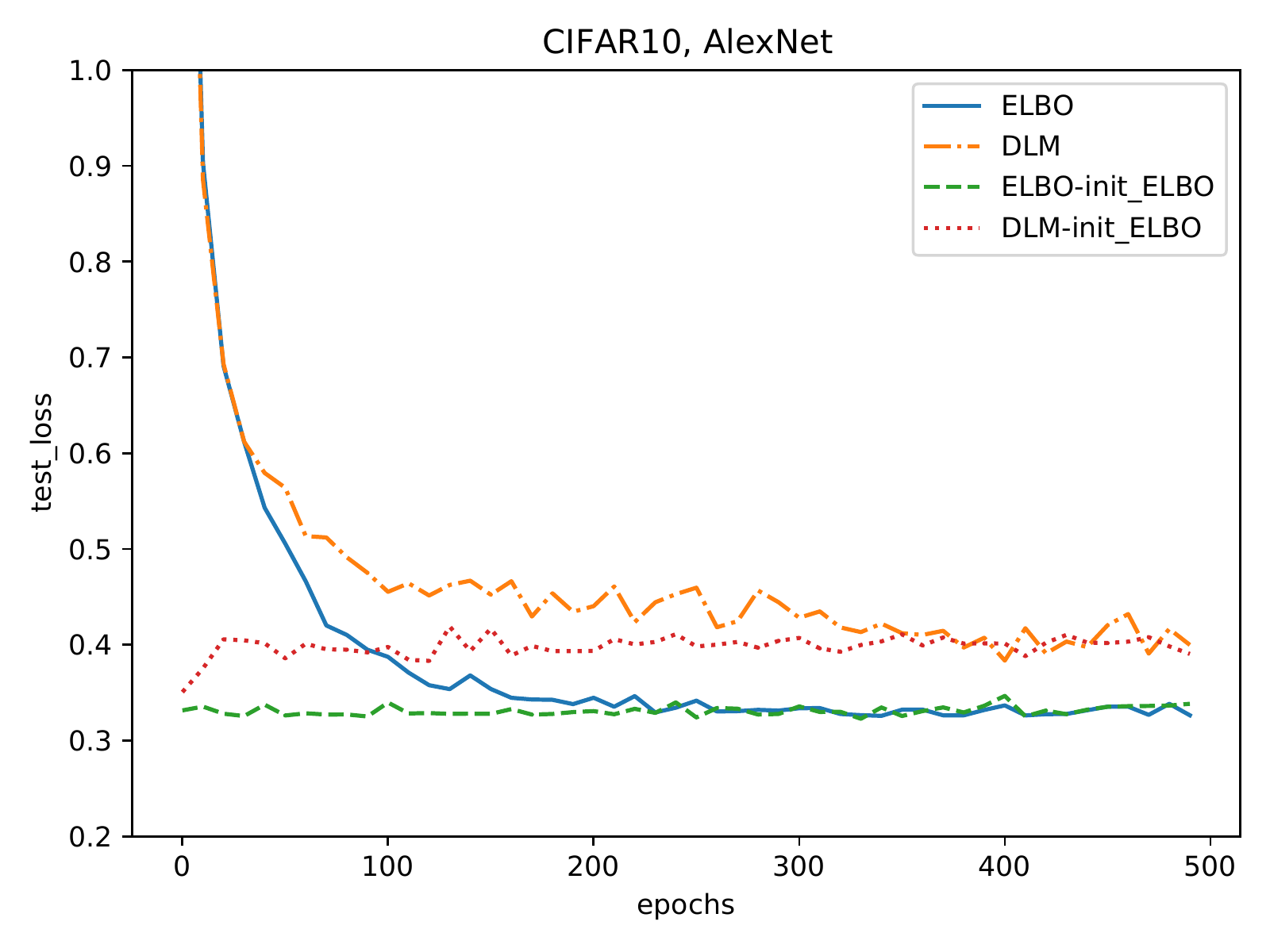}
         \caption{test loss($\downarrow$)}
         \label{fig:elbo_init}
    \end{subfigure}
    \begin{subfigure}[t]{0.32\textwidth}
         \centering
         \includegraphics[width=\textwidth]{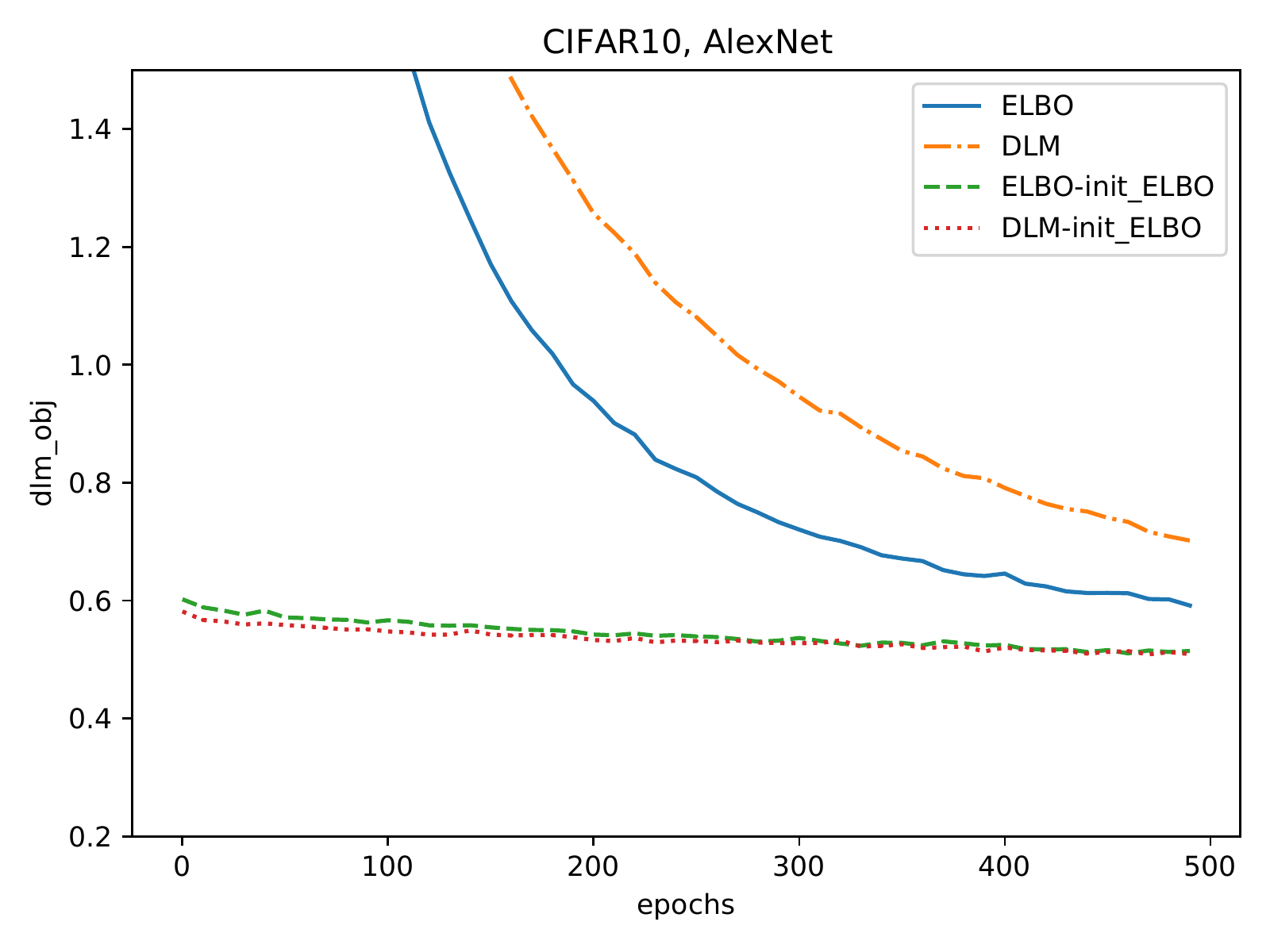}
         \caption{Trajectory of dlm loss}
         \label{fig:training-dlm-loss}
    \end{subfigure}
    \begin{subfigure}[t]{0.32\textwidth}
         \centering
         \includegraphics[width=\textwidth]{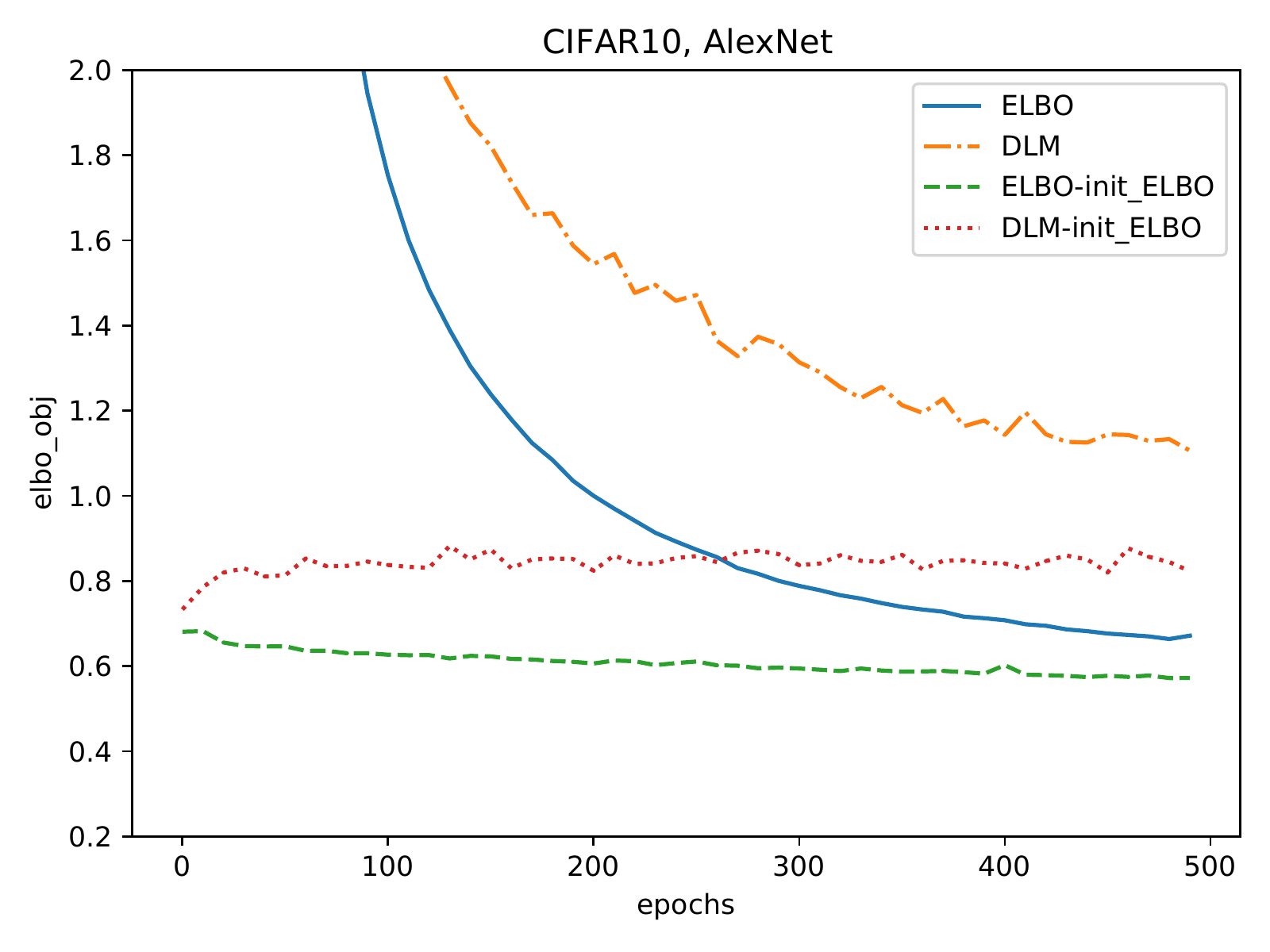}
         \caption{Trajectory of elbo loss}
         \label{fig:training-elbo-loss}
    \end{subfigure}
    \caption{Comparison of ELBO and DLM with/without initialization}
\end{figure}

\begin{observation}
\label{obs:elbo-better}
    ELBO performs better than DLM.
\end{observation}
DLM is worse than ELBO in most experiments.
A summary over all experiments is shown in Figure \ref{fig:all-result} and a concrete learning curve is shown in Figure \ref{fig:elbo_init}.
In a few cases DLM has similar performance but it does not outperform ELBO in any of the experiments. 
The range of test losses in the experiments is up to 2, so the differences shown are significant.

To better understand the behavior of ELBO and DLM, we also compute some quantities during the training, including elbo objective, dlm objective and KL divergence, and see how they change. We repeat with different seeds and the quantities behave similarly regardless of the random seed we choose.
To our surprise, we observe:
\begin{observation}
\label{obs:elbo-opt-dlm-better}
    ELBO appears to optimize the dlm objective better than the DLM algorithm.
\end{observation}
As shown in Figure \ref{fig:training-dlm-loss}, which depicts how the dlm loss changes during training, ELBO (blue solid line) is below DLM (orange dashdot line). At the same time, ELBO optimizes its own objective, elbo objective, better than DLM, as shown in Figure \ref{fig:training-elbo-loss}. One might suspect that the reason for this is that the dlm objective is likely to get stuck in local optima. Thus if we initialize DLM with a good starting point, then DLM may be improved.
To test this, we initialize DLM with ELBO solution and continue to train with dlm objective and we denote this solution as DLM-init\_ELBO. For comparison we also have ELBO-init\_ELBO which is initialized with ELBO and then continue to train with elbo objective. 
However, DLM is not improved with ELBO initialization and it even makes ELBO worse, as shown in Figure \ref{fig:elbo_init}. The increase of test loss for DLM shows that DLM is not stuck in local optima by accident, but is inherently worse than ELBO.
\begin{observation}
    The failure of DLM is not due to local optima.
\end{observation}

The good news is that with ELBO initialization, DLM optimizes the dlm objective slightly better than ELBO, as shown in Figure \ref{fig:training-dlm-loss}. We note that this does not happen for every experiment. In Figure \ref{fig:dlm-CIFAR100} in the \myappendix, DLM still optimizes dlm loss worse than ELBO even with ELBO initialization.

\begin{figure}
    \centering
    \begin{subfigure}[t]{0.32\textwidth}
         \centering
         \includegraphics[width=\textwidth]{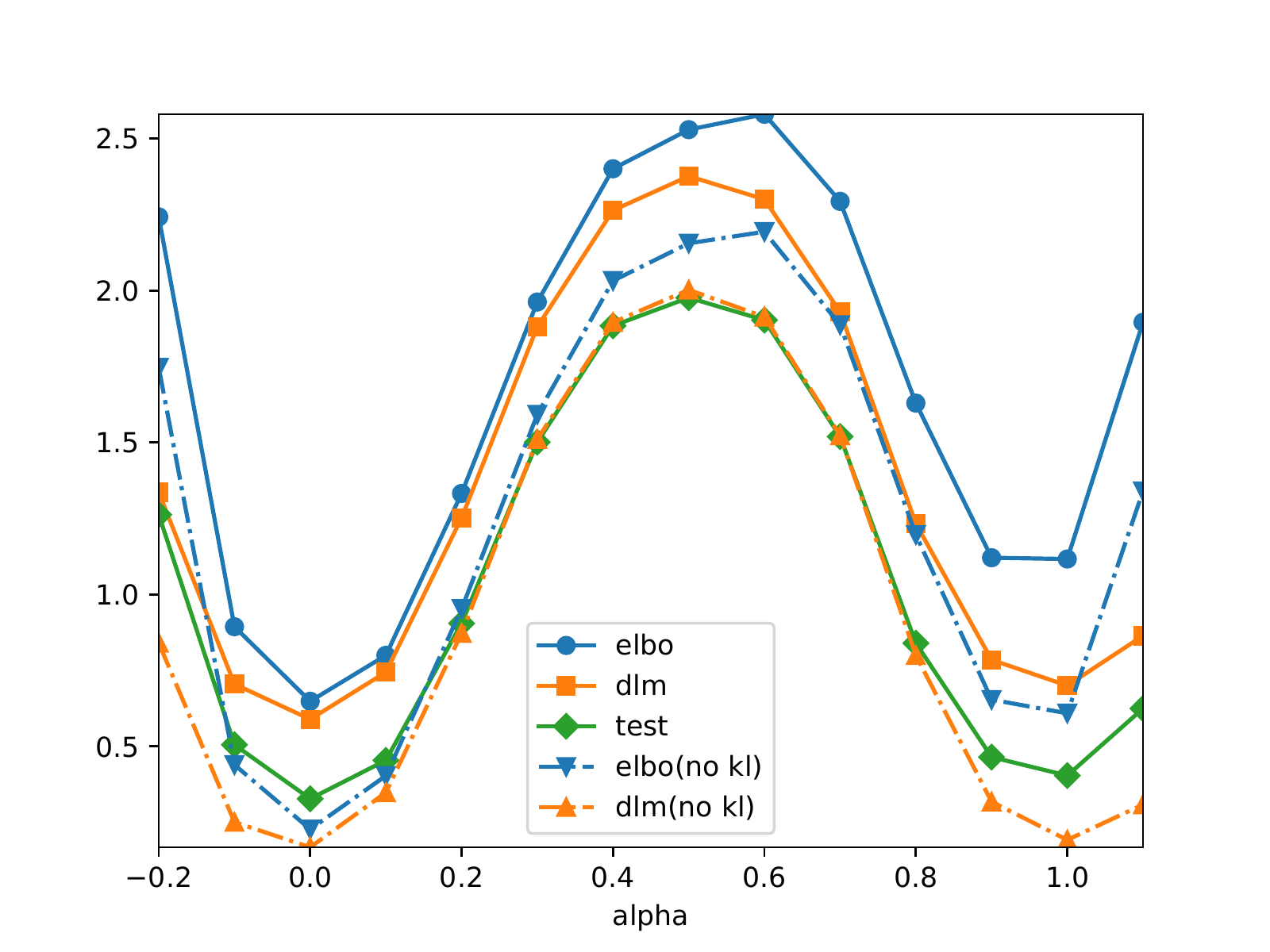}
         \caption{ELBO and DLM}
         \label{fig:no-init-surface}
    \end{subfigure}
    \begin{subfigure}[t]{0.32\textwidth}
         \centering
         \includegraphics[width=\textwidth]{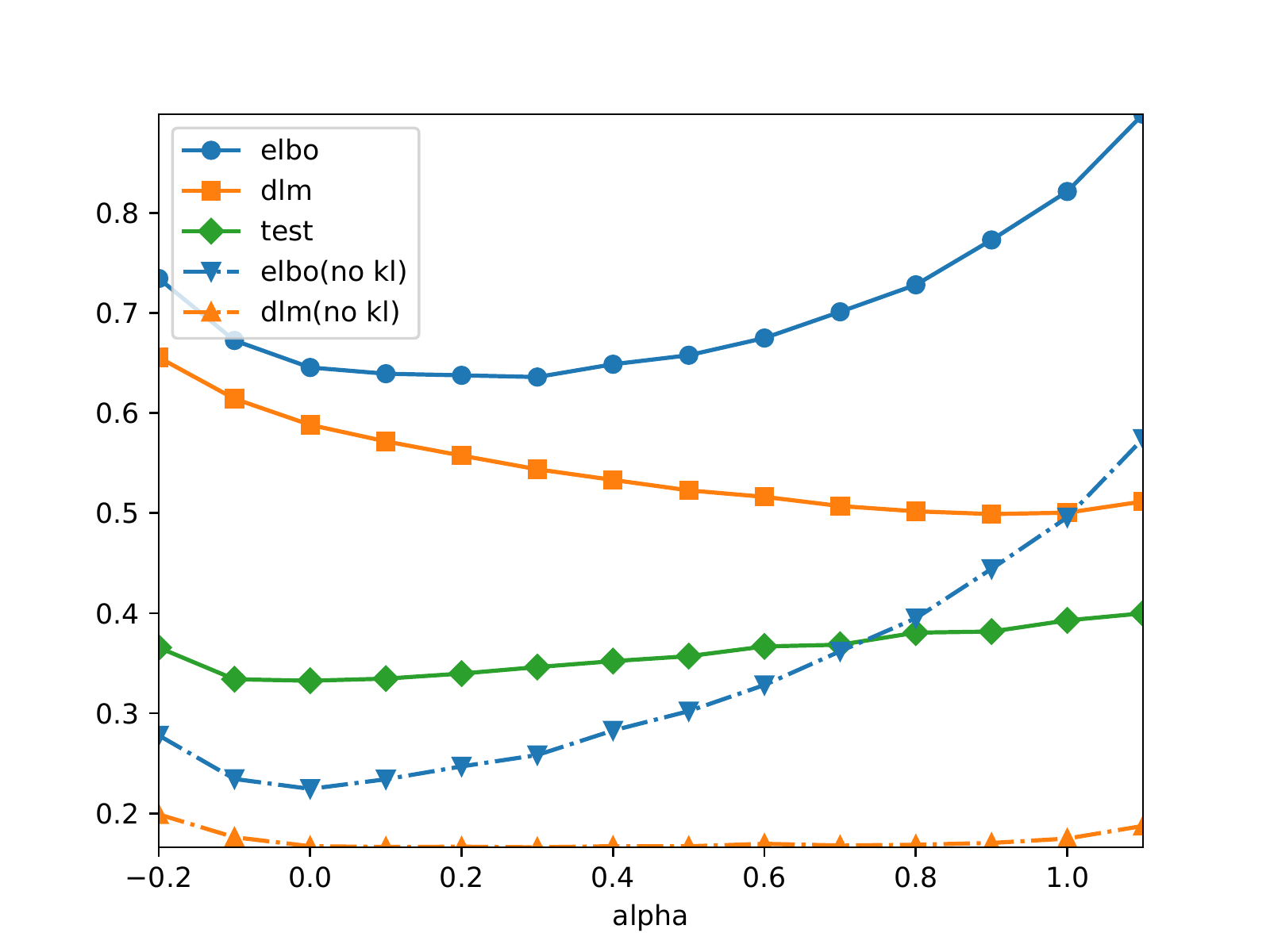}
         \caption{ELBO and DLM-init\_ELBO}
         \label{fig:mix-surface}
    \end{subfigure}
    \begin{subfigure}[t]{0.32\textwidth}
         \centering
         \includegraphics[width=\textwidth]{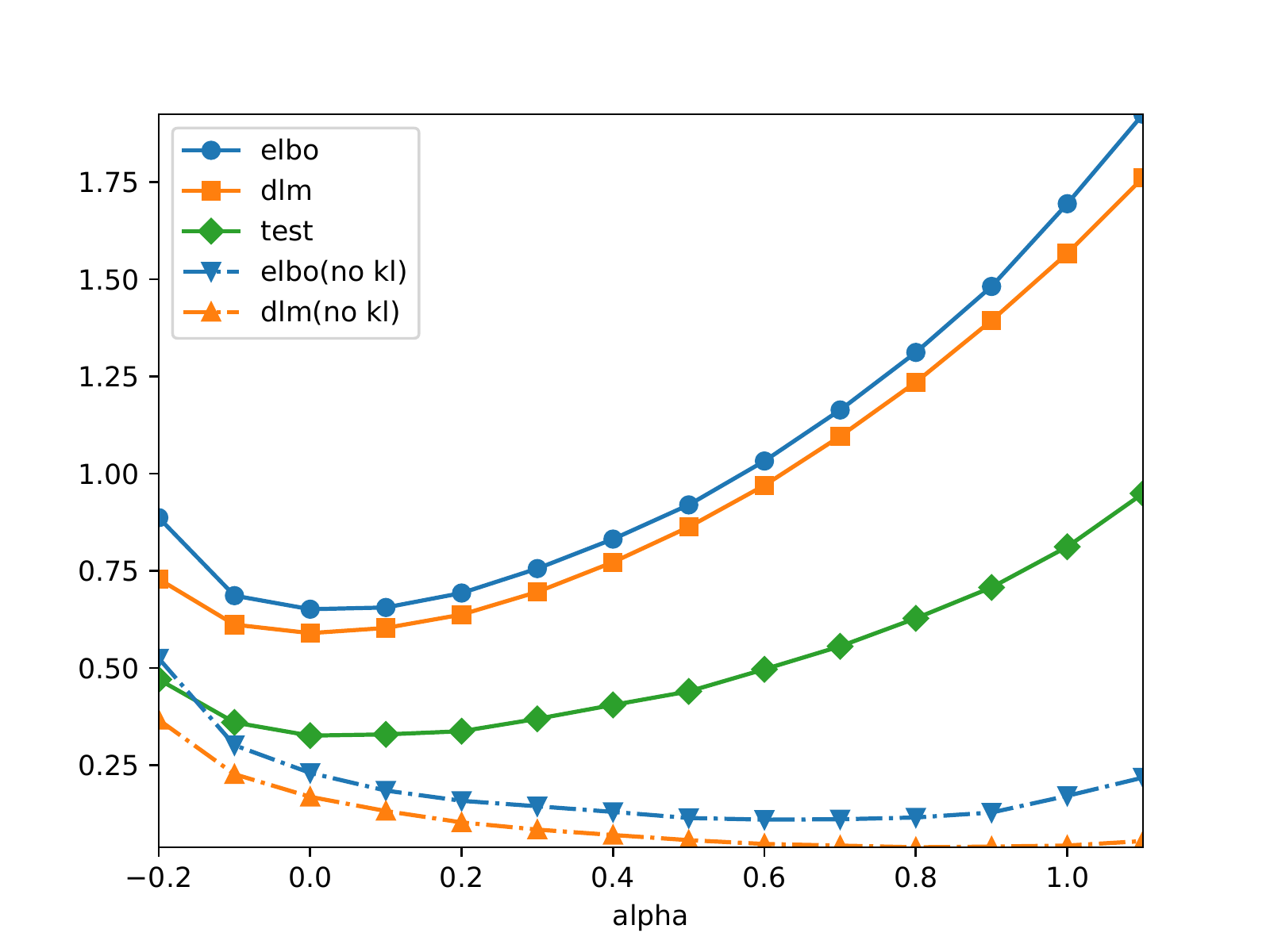}
         \caption{ELBO and DLM-init-no\_kl}
         \label{fig:surface-zero-mix}
    \end{subfigure}
    \caption{Loss Surface}
    \label{fig:all-surface}
\end{figure}

All these abnormalities lead us to further explore the structure of elbo and dlm losses. Motivated by \cite{loss-surface}, we create a path from ELBO to DLM, i.e. $\mu=(1-\alpha) \mu_{\elbo} + \alpha \mu_{\dlm}, \sigma^2 = (1-\alpha) \sigma^2_\elbo + \alpha \sigma^2_\dlm$, and then evaluate the elbo objective (with/without KL), dlm objective (with/without KL) and test loss on $(\mu, \sigma^2)$. It is clear that ELBO corresponds to $\alpha=0$ and DLM corresponds to $\alpha=1$. From the loss surface plotted in Figure \ref{fig:no-init-surface}, we confirm Observation \ref{obs:elbo-better} and \ref{obs:elbo-opt-dlm-better} (note that the dlm objective is lower at $\alpha=0$). Figure \ref{fig:mix-surface} plots the path from ELBO to DLM-init\_ELBO, and we can see how optimizing with dlm loss will change these loss functions. 
\begin{observation}
\label{obs:consistent}
    The elbo objective with $\eta=0.1$ is better aligned with test loss than the dlm objective, which indicates that the elbo objective generalizes better.
\end{observation}
Figure \ref{fig:mix-surface} shows that as $\alpha$ increases, the dlm objective decreases, but both the elbo objective and the test loss increase. 
Figure \ref{fig:mix-surface-cifar100} in the {\myappendix} shows a different situation (on CIFAR100 with AlexNet) where the dlm objective also increases, i.e., it is also aligned. But in our experiments the elbo objective and the test loss are always aligned in our experiments. Observation \ref{obs:consistent} also explains the abnormality in Figure \ref{fig:dlm-CIFAR100}, in which DLM-init\_ELBO significantly increases the dlm objective in first few epochs. This is because we optimize the dlm objective within a batch but plot the average dlm objective value among all batches. The poor generalization of the dlm objective may cause the value evaluated on other batches to increase and the overall value increases.

\begin{figure}
\vspace{-0.3cm}
    \centering
    \begin{subfigure}[t]{0.45\textwidth}
         \centering
         \includegraphics[width=\textwidth]{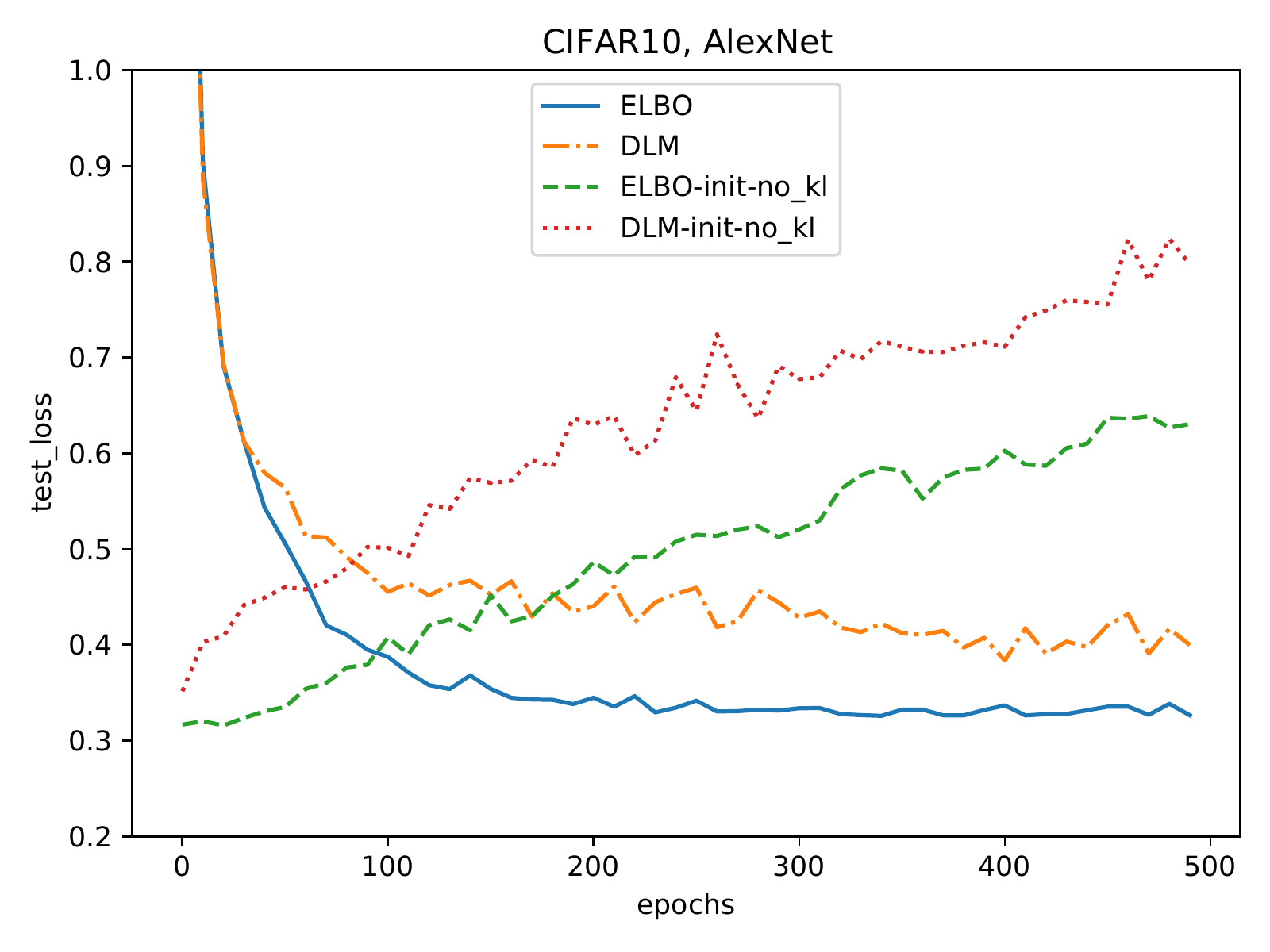}
         \caption{test loss}
         \label{fig:test-zero}
    \end{subfigure}
    \begin{subfigure}[t]{0.48\textwidth}
         \centering
         \includegraphics[width=\textwidth]{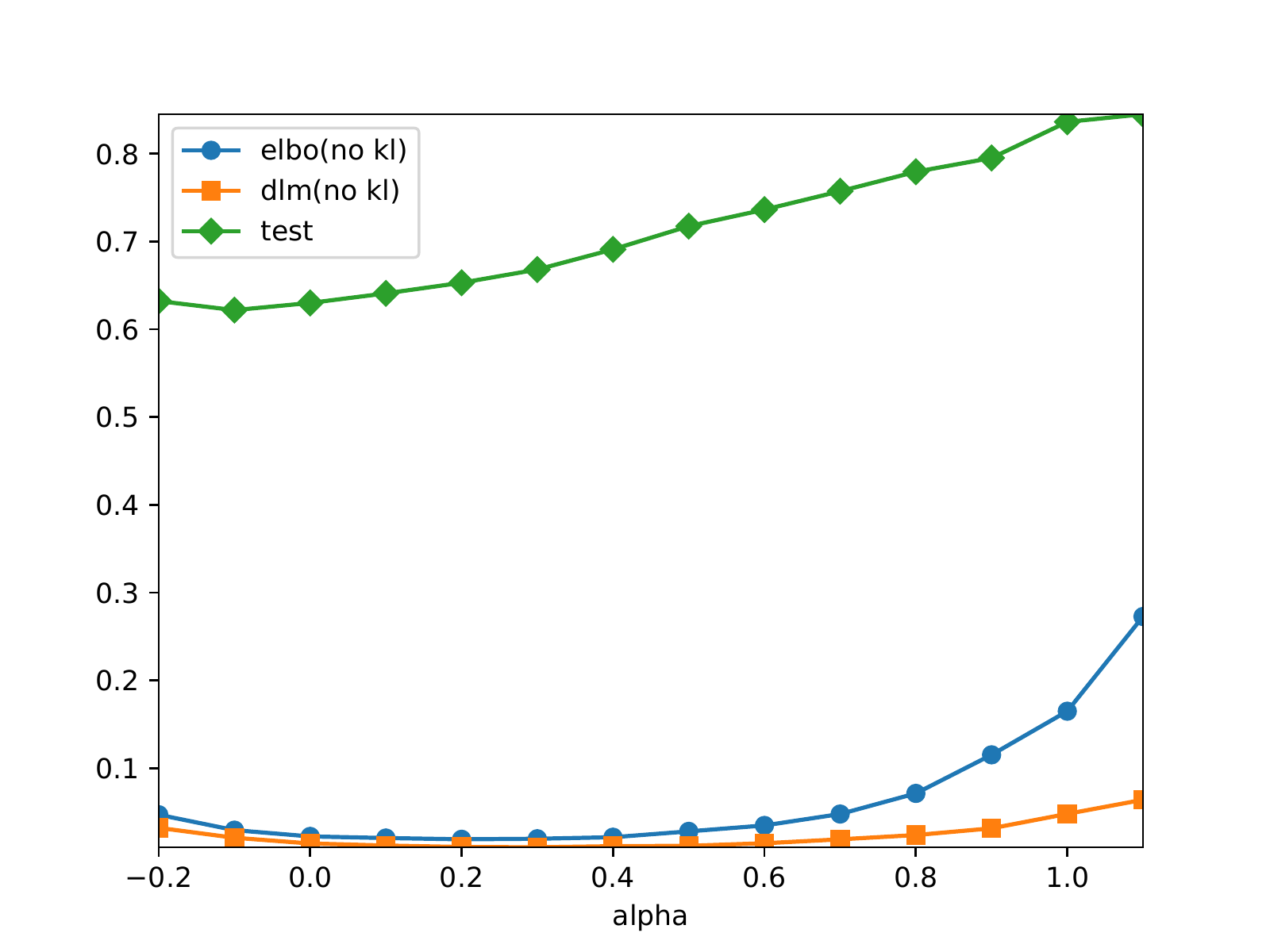}
         \caption{ELBO-init-no\_kl and DLM-init-no\_kl}
         \label{fig:surface-zero}
    \end{subfigure}
    \caption{Test Performance and Loss Surface for $\eta=0$}
\end{figure}

We also observe from Figure \ref{fig:mix-surface} that the dlm objective goes down but its loss term goes up, implying that the reduction in objective is due to the KL term. The same sensitivity regarding the {\em tradeoff between the loss term and regularizer} appears in other cases as well. To explore this we reduce $\eta$ to 0 after initializing with ELBO. Figure \ref{fig:test-zero} shows that reducing $\eta$ to 0 makes both ELBO and DLM perform worse than their original version but the relationship of ELBO-init-no\_kl and DLM-init-no\_kl still follows Observation \ref{obs:elbo-better}. Figure \ref{fig:surface-zero} again shows Observation \ref{obs:elbo-opt-dlm-better}, i.e., that the dlm objective ($\eta=0$) achieves lower value at ELBO-init-no\_kl than at DLM-init-no\_kl. 

In contrast with Figure \ref{fig:surface-zero}, Figure \ref{fig:surface-zero-mix} depicts the path between DLM-init-no\_kl and the original ELBO solution (which is the best), instead of ELBO-init-no\_kl. Then we can see that neither  the elbo loss nor the dlm loss without KL is aligned with the test loss, indicating overfitting. From another view, the three plots in Figure \ref{fig:all-surface} depict the change of loss functions along three directions from ELBO. The elbo objective with $\eta=0.1$ is aligned with the test loss in all three cases. But we cannot find such proper $\eta$ for dlm. In (a) and (c), the dlm objective with $\eta=0.1$ is aligned with the test loss, but in (b) the dlm objective with $\eta=0$ is aligned with the test loss. All these results support Observation \ref{obs:consistent}. 

In addition to the work mentioned above, we have explored bounded optimization, smoothed loss, collapsed variational inference \citep{collapsed-elbo} and empirical Bayes \citep{dvi}. The first two measures aim to close the gap between theoretical analysis and real applications so that we can utilize the upper bounds to guarantee the performance of DLM. The latter two define a hierarchical model and perform inference on the prior parameters, which results in a different regularizer to replace the original KL divergence. 
Although these measures can sometimes improve the performance of DLM, they do not help DLM outperform ELBO. 
Details are provided in the \myappendix.

Overall, we found that at least one of Observation \ref{obs:elbo-opt-dlm-better} and \ref{obs:consistent} appears in all experiments. In cases where ELBO does not optimize the dlm objective better than DLM, Observation \ref{obs:consistent} kicks in and shows that optimizing the dlm objective cannot make the performance better; In cases where the dlm objective is aligned with the test loss, we find that ELBO optimizes the dlm objective better. Thus, none of the variants of DLM mentioned in this paper outperforms ELBO.  

\section{Conclusion and Future Work}
Direct loss minimization has a strong motivation that we should use the same loss function in both training and testing. During training, we add a regularizer to prevent overfitting. Many theoretical results guarantee the performance of DLM optimizers. Despite its empirical success in sparse Gaussian processes, we observe that such success does not appear for Bayesian neural networks. In empirical exploration, we found that the goal of DLM is also severely challenged as ELBO optimizes dlm objective better than DLM itself. The most likely reason for this is that the dlm objective is hard to optimize for Bayeisan neural networks. 
Besides, DLM generalizes worse than ELBO, because elbo loss is more consistent with test loss, pointing out overfitting of the dlm objective. This relates to data misspecification as suggested in \citep{pac-m} but how to test the notion of misspecification in image classification remains unclear as the neural networks are expressive. 
It would be interesting to explore what distinguishes cases where DLM succeeds, such as sparse Gaussian processes, from the behavior shown in this paper.
As mentioned above, we can view the elbo loss as a (potentially better behaved) surrogate loss of the true loss given by dlm. It would be interesting to explore theoretical analysis that explains differences in behavior from this perspective. 


\begin{ack}
This work was partly supported by NSF under grant IIS-1906694. Some of the experiments in this paper were run on the Big Red computing system at Indiana University, supported in part by Lilly Endowment, Inc., through its support for the Indiana University Pervasive Technology Institute.
\end{ack}

\bibliography{local}

\begin{thebibliography}{13}
\providecommand{\natexlab}[1]{#1}
\providecommand{\url}[1]{\texttt{#1}}
\expandafter\ifx\csname urlstyle\endcsname\relax
  \providecommand{\doi}[1]{doi: #1}\else
  \providecommand{\doi}{doi: \begingroup \urlstyle{rm}\Url}\fi

\bibitem[Dusenberry et~al.(2020)Dusenberry, Jerfel, Wen, Ma, Snoek, Heller,
  Lakshminarayanan, and Tran]{bnn-rank1}
Michael~W. Dusenberry, Ghassen Jerfel, Yeming Wen, Yi-An Ma, Jasper Snoek,
  Katherine Heller, Balaji Lakshminarayanan, and Dustin Tran.
\newblock Efficient and scalable bayesian neural nets with rank-1 factors.
\newblock In \emph{Proceedings of the 37th International Conference on Machine
  Learning}, ICML'20. JMLR.org, 2020.

\bibitem[Garipov et~al.(2018)Garipov, Izmailov, Podoprikhin, Vetrov, and
  Wilson]{loss-surface}
Timur Garipov, Pavel Izmailov, Dmitrii Podoprikhin, Dmitry Vetrov, and
  Andrew~Gordon Wilson.
\newblock Loss surfaces, mode connectivity, and fast ensembling of dnns.
\newblock In \emph{Proceedings of the 32nd International Conference on Neural
  Information Processing Systems}, NIPS'18, page 8803–8812, Red Hook, NY,
  USA, 2018. Curran Associates Inc.

\bibitem[He et~al.(2016)He, Zhang, Ren, and Sun]{preresnet}
Kaiming He, Xiangyu Zhang, Shaoqing Ren, and Jian Sun.
\newblock Identity mappings in deep residual networks.
\newblock In Bastian Leibe, Jiri Matas, Nicu Sebe, and Max Welling, editors,
  \emph{Computer Vision -- ECCV 2016}, pages 630--645, Cham, 2016. Springer
  International Publishing.
\newblock ISBN 978-3-319-46493-0.

\bibitem[Jankowiak et~al.(2020)Jankowiak, Pleiss, and
  Gardner]{Jankowiak2020ParametricGP}
Martin Jankowiak, Geoff Pleiss, and Jacob~R. Gardner.
\newblock Parametric gaussian process regressors.
\newblock In \emph{ICML}, 2020.

\bibitem[Krizhevsky et~al.(2012)Krizhevsky, Sutskever, and Hinton]{alexnet}
Alex Krizhevsky, Ilya Sutskever, and Geoffrey~E Hinton.
\newblock Imagenet classification with deep convolutional neural networks.
\newblock In F.~Pereira, C.J. Burges, L.~Bottou, and K.Q. Weinberger, editors,
  \emph{Advances in Neural Information Processing Systems}, volume~25. Curran
  Associates, Inc., 2012.
\newblock URL
  \url{https://proceedings.neurips.cc/paper/2012/file/c399862d3b9d6b76c8436e924a68c45b-Paper.pdf}.

\bibitem[Meir and Zhang(2003)]{Meir-Zhang}
Ron Meir and Tong Zhang.
\newblock Generalization error bounds for bayesian mixture algorithms.
\newblock \emph{J. Mach. Learn. Res.}, 4\penalty0 (null):\penalty0 839–860,
  dec 2003.
\newblock ISSN 1532-4435.

\bibitem[Morningstar et~al.(2022)Morningstar, Alemi, and Dillon]{pac-m}
Warren~R. Morningstar, Alex Alemi, and Joshua~V. Dillon.
\newblock Pacm-bayes: Narrowing the empirical risk gap in the misspecified
  bayesian regime.
\newblock In Gustau Camps-Valls, Francisco J.~R. Ruiz, and Isabel Valera,
  editors, \emph{Proceedings of The 25th International Conference on Artificial
  Intelligence and Statistics}, volume 151 of \emph{Proceedings of Machine
  Learning Research}, pages 8270--8298. PMLR, 28--30 Mar 2022.
\newblock URL \url{https://proceedings.mlr.press/v151/morningstar22a.html}.

\bibitem[Sheth and Khardon(2017)]{sheth2017excess}
Rishit Sheth and Roni Khardon.
\newblock Excess risk bounds for the {B}ayes risk using variational inference
  in latent {G}aussian models.
\newblock In \emph{{NIPS}}, pages 5151--5161, 2017.

\bibitem[Sheth and Khardon(2019)]{Sheth2019pseudo}
Rishit Sheth and Roni Khardon.
\newblock Pseudo-bayesian learning via direct loss minimization with
  applications to sparse gaussian process models.
\newblock In \emph{Symposium on Advances in Approximate Bayesian Inference
  (AABI)}, 2019.

\bibitem[Tomczak et~al.(2021)Tomczak, Swaroop, Foong, and
  Turner]{collapsed-elbo}
Marcin~B. Tomczak, Siddharth Swaroop, Andrew Y.~K. Foong, and Richard~E Turner.
\newblock Collapsed variational bounds for bayesian neural networks.
\newblock In A.~Beygelzimer, Y.~Dauphin, P.~Liang, and J.~Wortman Vaughan,
  editors, \emph{Advances in Neural Information Processing Systems}, 2021.
\newblock URL \url{https://openreview.net/forum?id=ykN3tbJ0qmX}.

\bibitem[Wei et~al.(2021)Wei, Sheth, and Khardon]{dlm-sgp}
Yadi Wei, Rishit Sheth, and Roni Khardon.
\newblock Direct loss minimization for sparse gaussian processes.
\newblock In Arindam Banerjee and Kenji Fukumizu, editors, \emph{Proceedings of
  The 24th International Conference on Artificial Intelligence and Statistics},
  volume 130 of \emph{Proceedings of Machine Learning Research}, pages
  2566--2574. PMLR, 13--15 Apr 2021.
\newblock URL \url{https://proceedings.mlr.press/v130/wei21b.html}.

\bibitem[Wilson et~al.(2022)Wilson, Izmailov, Hoffman, Gal, Li, Pradier,
  Vikram, Foong, Lotfi, and Farquhar]{competition-approximate-inference}
Andrew~Gordon Wilson, Pavel Izmailov, Matthew~D Hoffman, Yarin Gal, Yingzhen
  Li, Melanie~F Pradier, Sharad Vikram, Andrew Foong, Sanae Lotfi, and
  Sebastian Farquhar.
\newblock Evaluating approximate inference in bayesian deep learning.
\newblock In Douwe Kiela, Marco Ciccone, and Barbara Caputo, editors,
  \emph{Proceedings of the NeurIPS 2021 Competitions and Demonstrations Track},
  volume 176 of \emph{Proceedings of Machine Learning Research}, pages
  113--124. PMLR, 06--14 Dec 2022.
\newblock URL \url{https://proceedings.mlr.press/v176/wilson22a.html}.

\bibitem[Wu et~al.(2019)Wu, Nowozin, Meeds, Turner, Hernandez-Lobato, and
  Gaunt]{dvi}
Anqi Wu, Sebastian Nowozin, Edward Meeds, Richard~E. Turner, Jose~Miguel
  Hernandez-Lobato, and Alexander~L. Gaunt.
\newblock Deterministic variational inference for robust bayesian neural
  networks.
\newblock In \emph{International Conference on Learning Representations}, 2019.
\newblock URL \url{https://openreview.net/forum?id=B1l08oAct7}.

\end{thebibliography}
\bibliographystyle{plainnat}


\newpage
\appendix

\counterwithin{figure}{section}

\section{Proof of Proposition 1}
\begin{proof}
We prove the claim by contradiction. If there exists $q' \in \set Q_{A_\eta}$ such that $\sum_{(x,y)\in D}-\log \E_{q'(\theta)}[p(y|\theta, x)]< \sum_{(x,y)\in D}-\log \E_{q_{\dlm}^{(\eta)} (\theta)}[p(y|\theta, x)]$. Then
\begin{align*}
    \sum_{(x, y)\in D}-\log \E_{q'(\theta)}[p(y|\theta, x)] + \eta \kl(q'||p) &\leq \sum_{(x, y)\in D} -\log \E_{q'(\theta)} [p(y|\theta, x)] + \eta A_\eta \\
    &< \sum_{(x,y)\in D}-\log \E_{q_{\dlm}^{(\eta)} (\theta)}[p(y|\theta, x)] + \eta \kl(q_{\dlm}^{(\eta)} || p).
\end{align*}
This contradicts the fact that $q_{\dlm}^{(\eta)}$ minimizes the log-loss plus regularizer. 
Thus, 
$q_{\dlm}^{(\eta)}=q_{\erm}^{(A_\eta)}$. 
\end{proof}

\section{Discussion}
\subsection{Bias in DLM}
One may suspect that the bias in estimating the loss term causes DLM to perform worse than ELBO. 
We believe that this is not the case. First, the same issue with biased gradient estimates exists in sparse Gaussian processes where both theoretical analysis and empirical results show that DLM is successful \citep{dlm-sgp}. 
Second. 
notice that in Bayesian neural networks, it is impossible to compute the log-loss exactly in evaluation. So the test loss is also biased and DLM is targeted for optimizing this biased loss. See related discussion by \citet{Jankowiak2020ParametricGP}.

\subsection{Choice of the Regularization Parameter}
Our comparisons in the main paper focus on fixing the regularization parameter $\eta$ to be 0.1. Notice that the elbo loss is an upper bound for the dlm loss. Thus, using the same $\eta$ means a stronger regularization in DLM.
This can potentially cause the regularization for DLM to be too strong, which might be another potential explanation for the bad performance of DLM. However, from Figure \ref{fig:vibo}, in which we perform bounded optimization with different bounds on the norm of the parameters, we observe that whichever bound we choose, DLM performs worse than ELBO. Recall that $\eta$ and the bound $A$ are related as specified in Proposition 1. If it is hard to find a bound that makes DLM perform better than ELBO, then it would be hard to find such $\eta$ that makes DLM perform better than ELBO. So the bad performance of DLM is not due to the choice of the regularization parameter $\eta$.

\section{Additional Results}
\subsection{ELBO initialization on CIFAR100 using AlexNet}
\begin{figure}[H]
    \centering
    \begin{subfigure}[t]{0.32\textwidth}
         \centering
         \includegraphics[width=\textwidth]{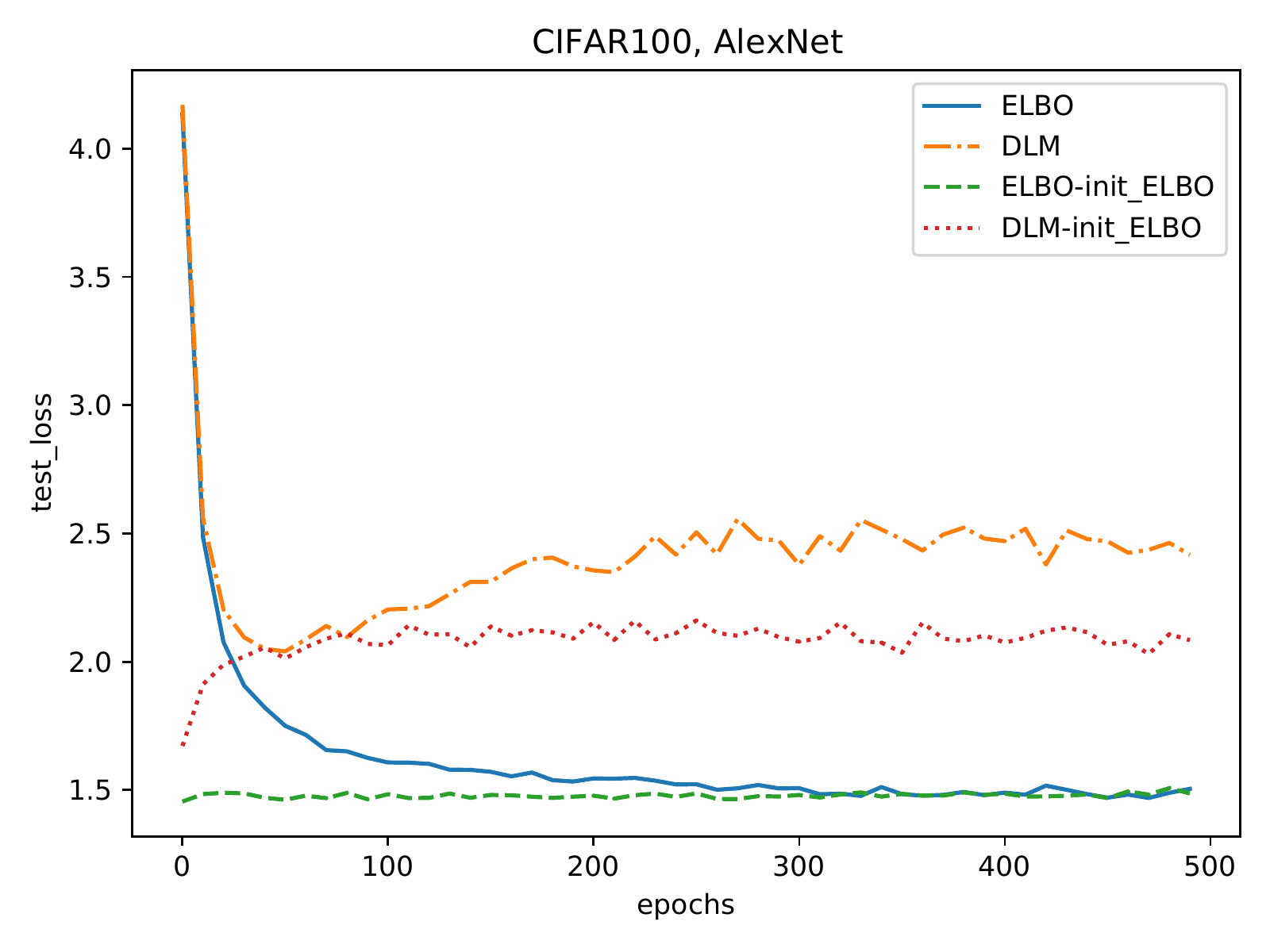}
         \caption{test loss($\downarrow$)}
    \end{subfigure}
    \begin{subfigure}[t]{0.32\textwidth}
         \centering
         \includegraphics[width=\textwidth]{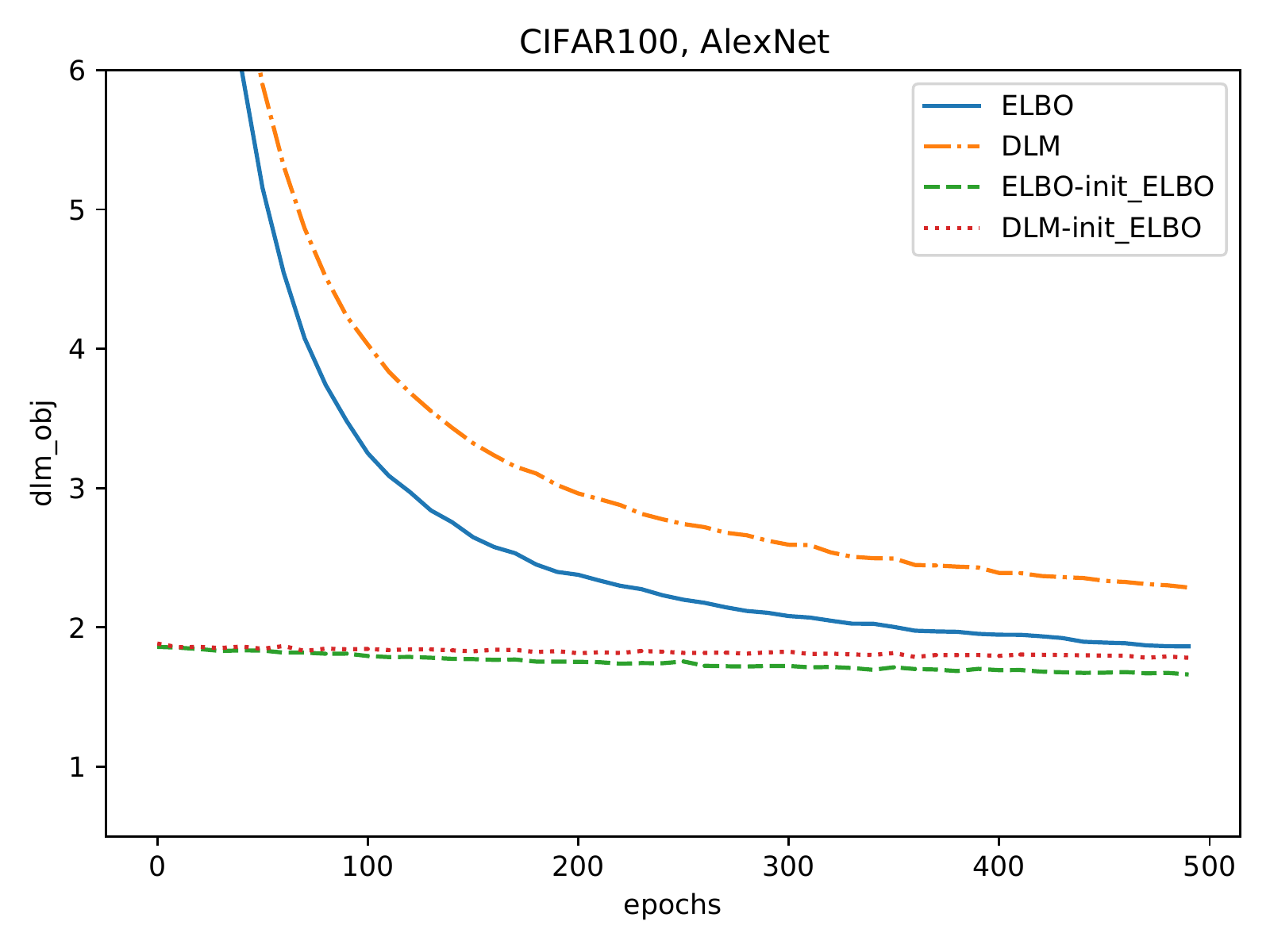}
         \caption{Trajectory of dlm loss}
         \label{fig:dlm-CIFAR100}
    \end{subfigure}
    \begin{subfigure}[t]{0.32\textwidth}
         \centering
         \includegraphics[width=\textwidth]{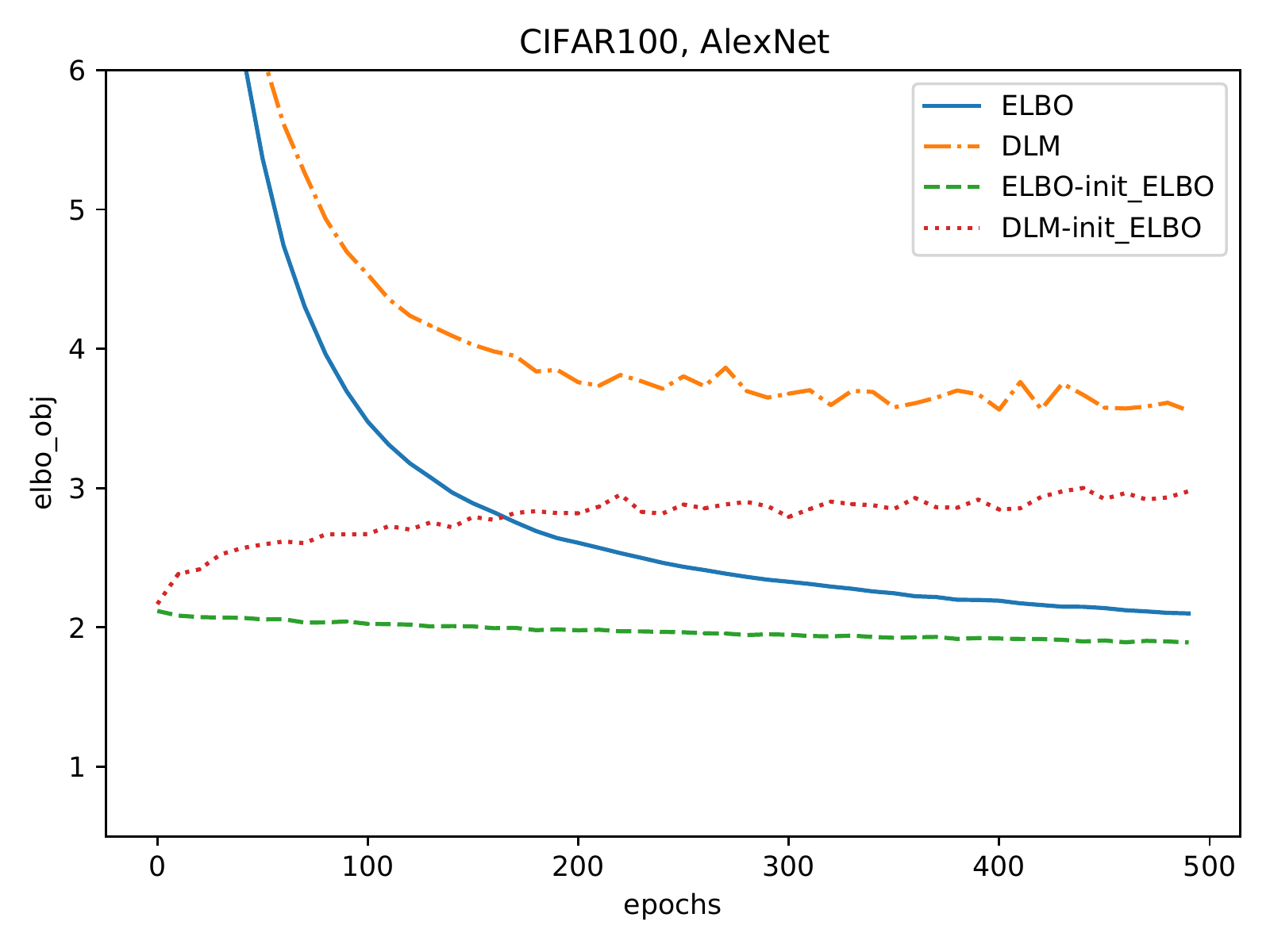}
         \caption{Trajectory of elbo loss}
         \label{fig:dlm-CIFAR100}
    \end{subfigure}
    \caption{Comparison of ELBO and DLM with/without initialization, on CIFAR100 with AlexNet}
    \label{fig:elbo-init-cifar100}
\end{figure}

\begin{figure}[H]
    \centering
    \begin{subfigure}[t]{0.48\textwidth}
         \centering
         \includegraphics[width=\textwidth]{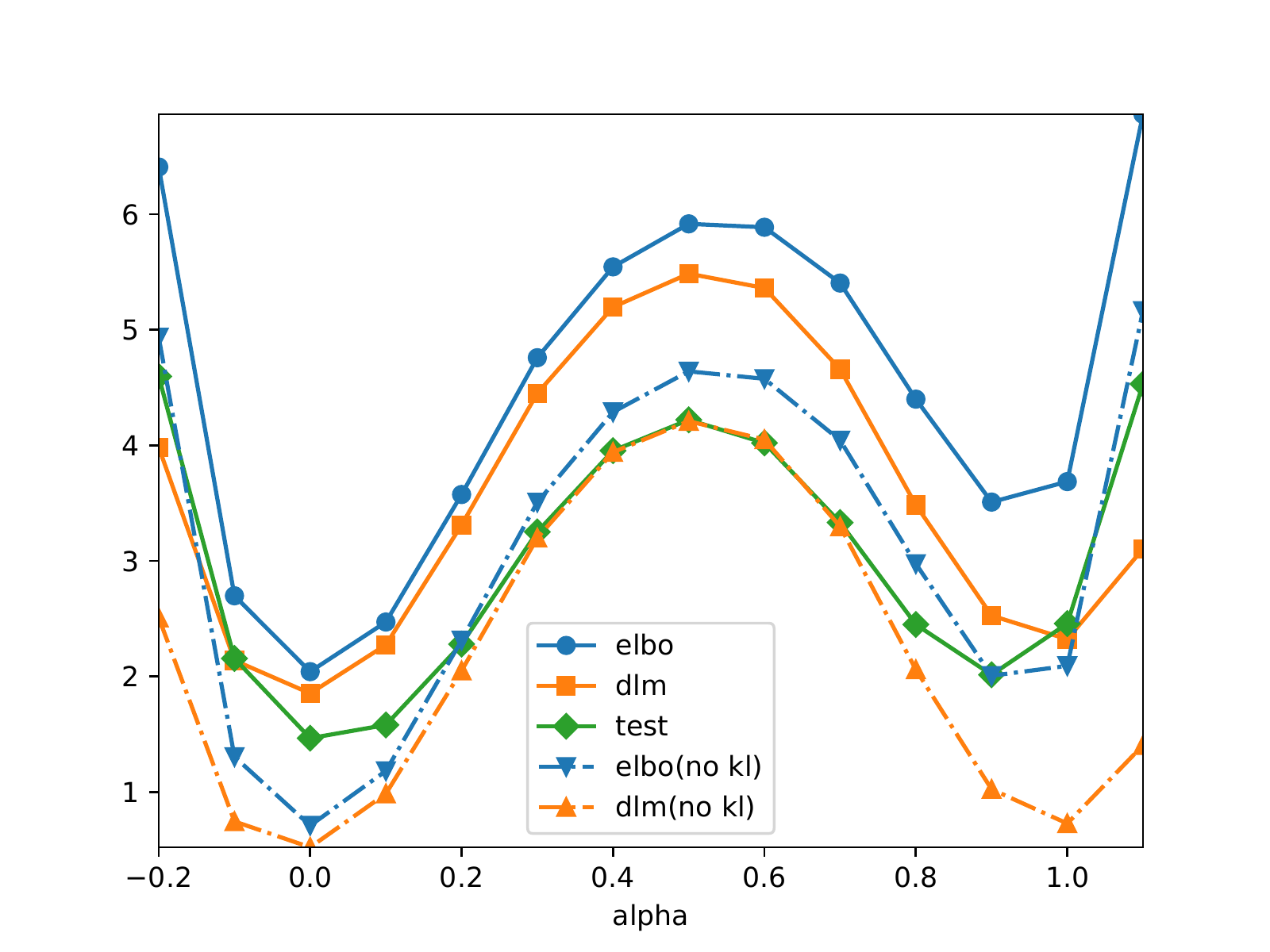}
         \caption{ELBO and DLM}
    \end{subfigure}
    \begin{subfigure}[t]{0.48\textwidth}
         \centering
         \includegraphics[width=\textwidth]{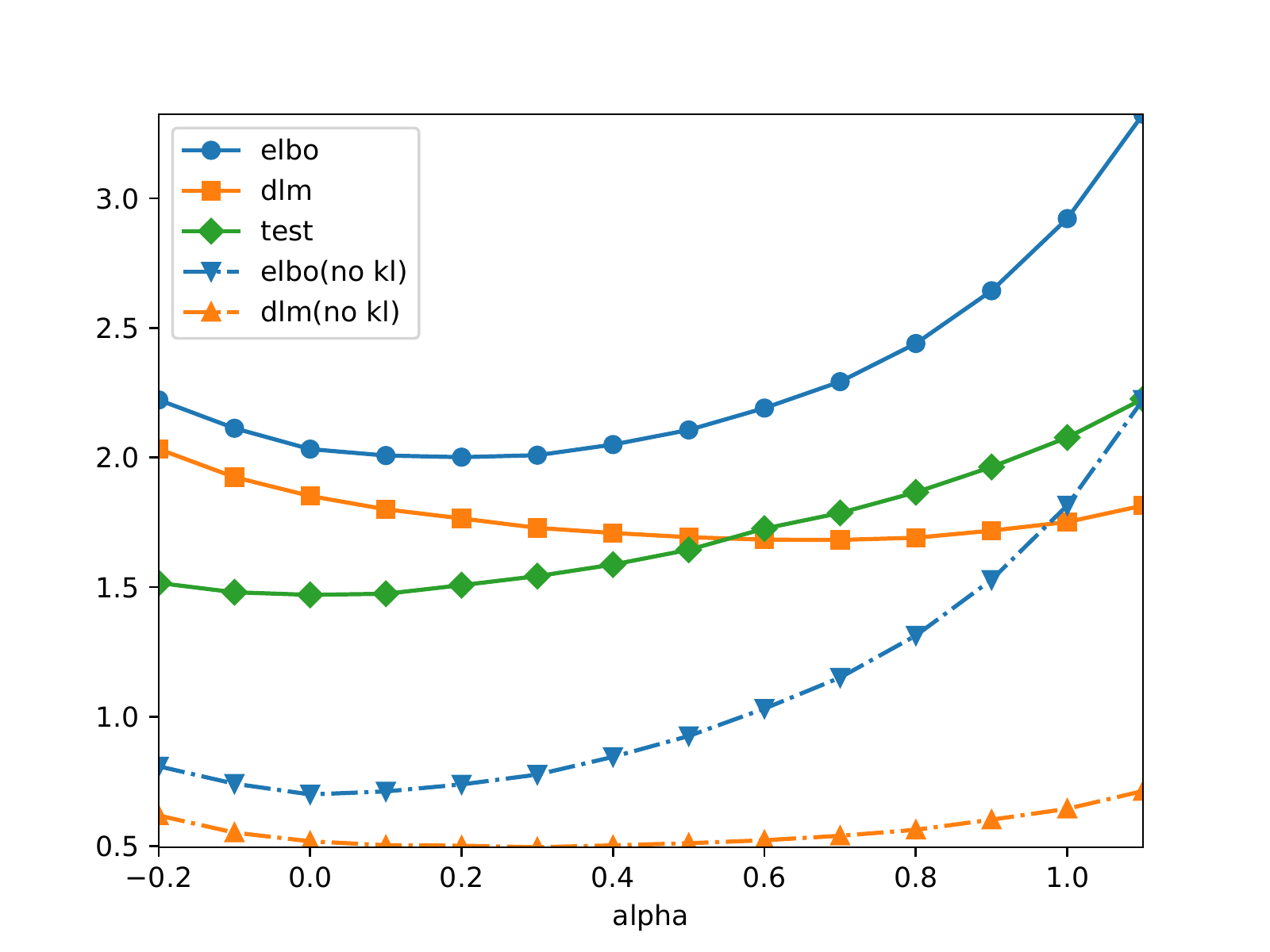}
         \caption{ELBO and DLM-init\_ELBO}
         \label{fig:mix-surface-cifar100}
    \end{subfigure}
    \caption{Loss Surface, on CIFAR100 with AlexNet}
    \label{fig:loss-surface-cifar100}
\end{figure}
Results for AlexNet on CIFAR100 are shown in Figure \ref{fig:elbo-init-cifar100} and \ref{fig:loss-surface-cifar100}. 
The behavior of ELBO and DLM on CIFAR100 is similar to that on CIFAR10, except in Figure \ref{fig:mix-surface-cifar100}, where the dlm loss is also consistent with test loss. Experiments on CIFAR100 do not break the two major observations we make in Section \ref{dlm-bnn}. 

\subsection{Smoothed Loss}
\label{smooth}
\begin{figure}[H]
    \centering
    \begin{subfigure}[t]{0.23\textwidth}
         \centering
         \includegraphics[width=\textwidth]{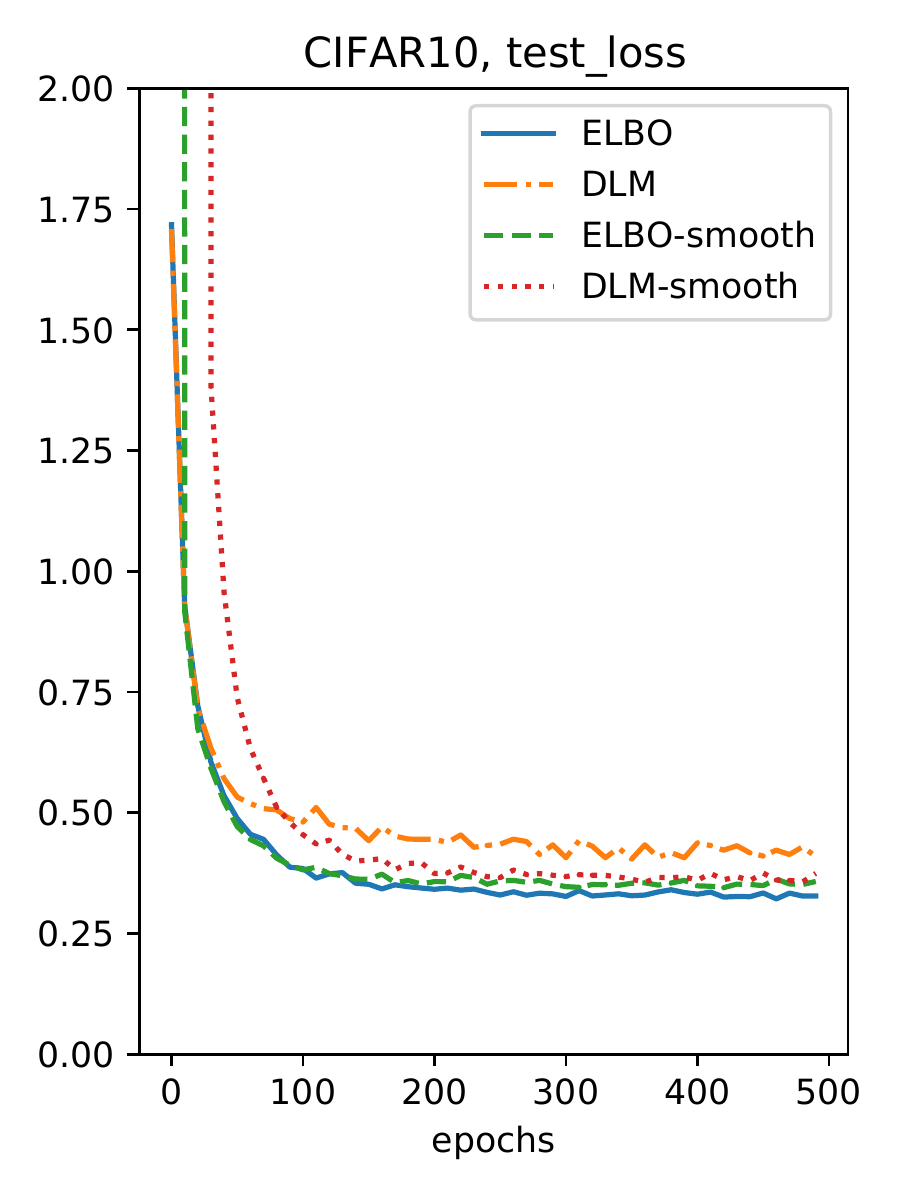}
         \caption{test loss, CIFAR10}
    \end{subfigure}
    \begin{subfigure}[t]{0.23\textwidth}
         \centering
         \includegraphics[width=\textwidth]{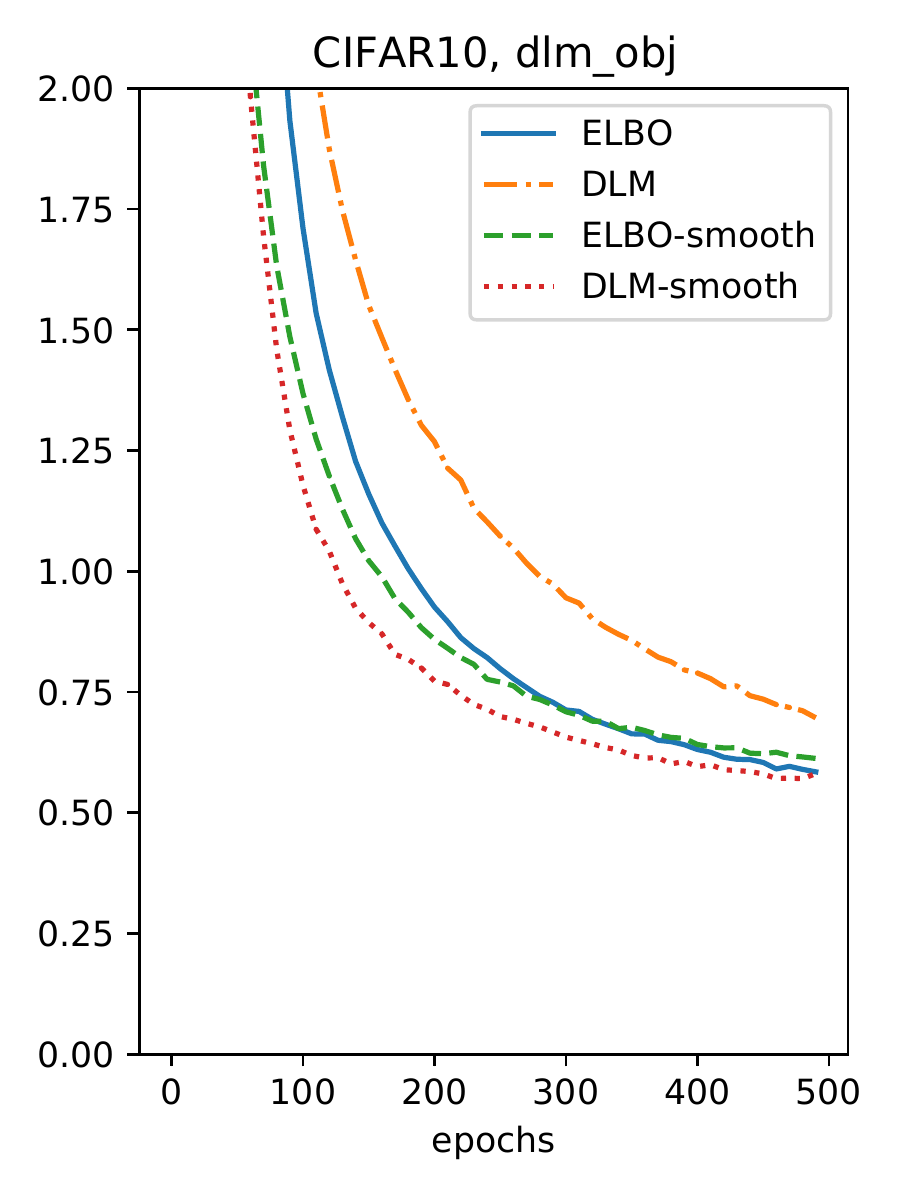}
         \caption{dlm loss, CIFAR10}
         \label{fig:smooth-dlm-cifar10}
    \end{subfigure}
    \begin{subfigure}[t]{0.23\textwidth}
         \centering
         \includegraphics[width=\textwidth]{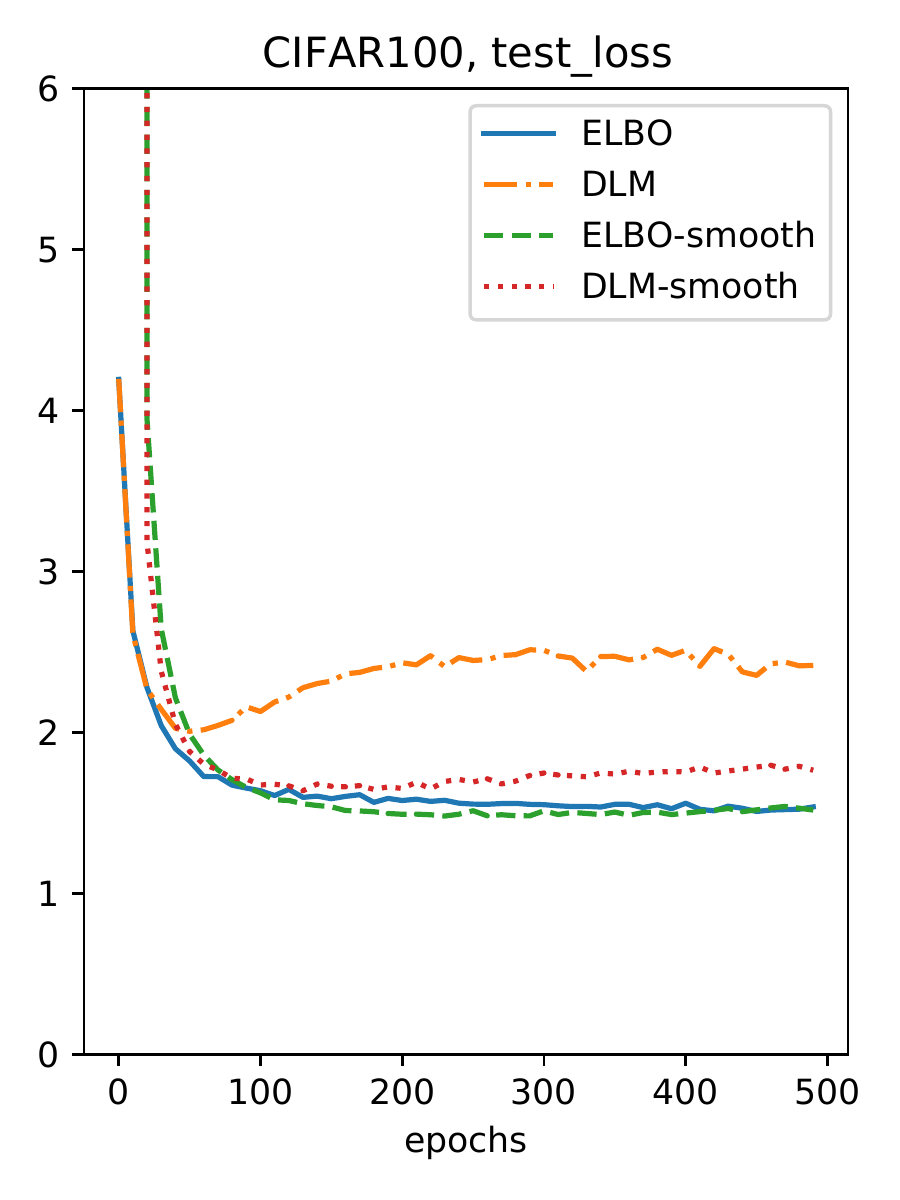}
         \caption{test loss, CIFAR100}
    \end{subfigure}
    \begin{subfigure}[t]{0.23\textwidth}
         \centering
         \includegraphics[width=\textwidth]{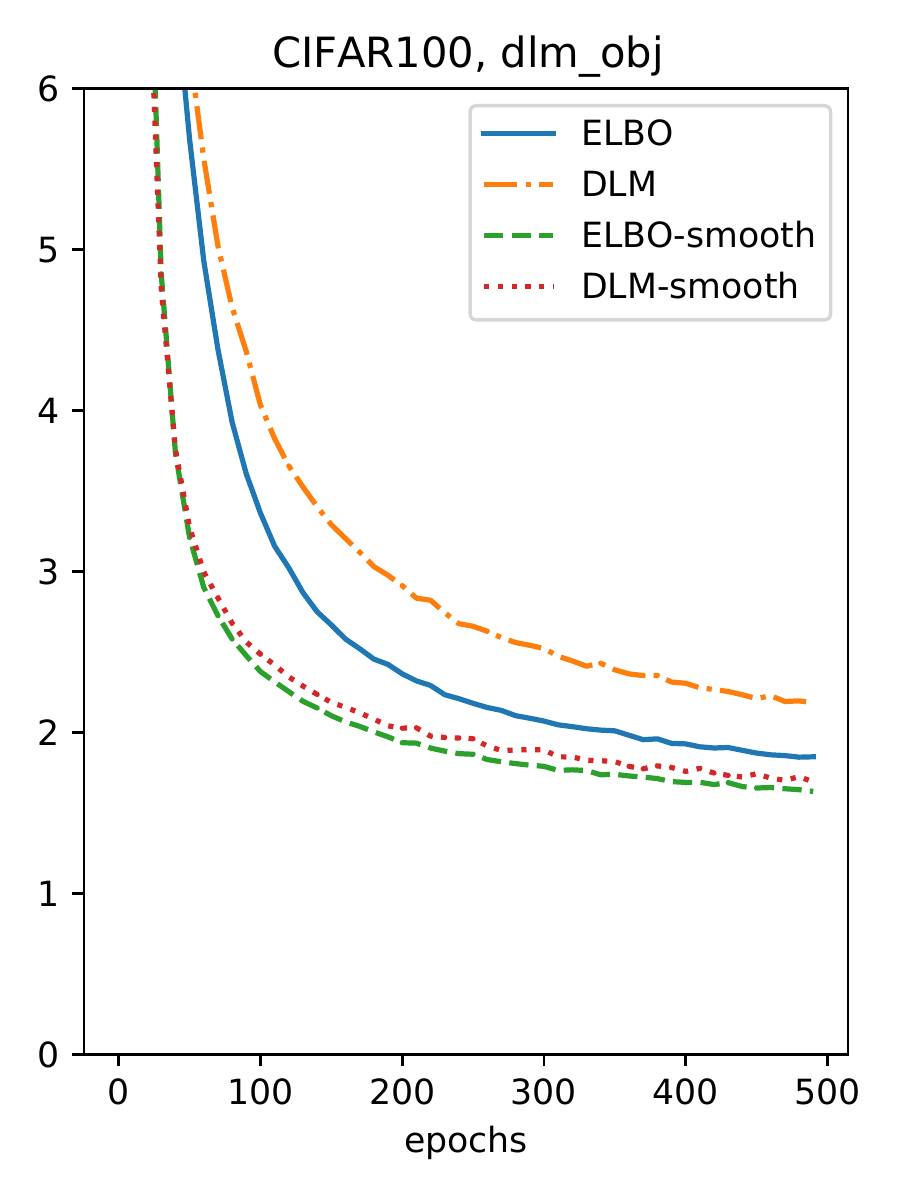}
         \caption{dlm loss, CIFAR100}
         \label{fig:smooth-dlm-cifar100}
    \end{subfigure}
    \caption{Comparison on losses($\downarrow$) of ELBO and DLM with/without smoothing}
    \label{fig:smooth}
\end{figure}

We use the smoothing factor $a=0.001$ in $\log^{(a)} p = \log((1-a)p + a)$. We use the smoothed loss only in elbo loss and dlm loss in training not in testing. Figure \ref{fig:smooth} shows that adding a smoothing factor does improve the performance of DLM, as in (a) and (c) DLM-smooth achieves lower test loss than DLM. However, DLM-smooth is still worse than ELBO. In Figure \ref{fig:smooth-dlm-cifar10} and \ref{fig:smooth-dlm-cifar100} we plot the dlm objectives during training. 
We observe that in Figure \ref{fig:smooth-dlm-cifar10}, ELBO does not optimize dlm loss better than DLM. However, since elbo loss generalizes better, DLM still perform worse than ELBO. This may indicate that the generalizability plays a more important role in explaining the failure of DLM. 

\subsection{Bounded Optimization}
It is hard to directly bound the KL divergence, so we instead bound the parameters as $\lVert \mu \rVert_2 \leq b_m$ and $\lVert \sigma^2 \rVert_{\infty} \leq b_v$. By analyzing the original solution of ELBO and DLM, we found $\lVert \mu_{\elbo} \rVert_2 $ and $\lVert \mu_{\dlm} \rVert_2$ are around 80 to 90 and $\lVert \sigma^2 \rVert_\infty$ is around 0.25. So we fix $b_v = 0.25$ and choose $b_m \in \{30, 50, 90, 120\}$. 
\begin{figure}[H]
    \centering
    \begin{subfigure}[t]{0.45\textwidth}
         \centering
         \includegraphics[width=\textwidth]{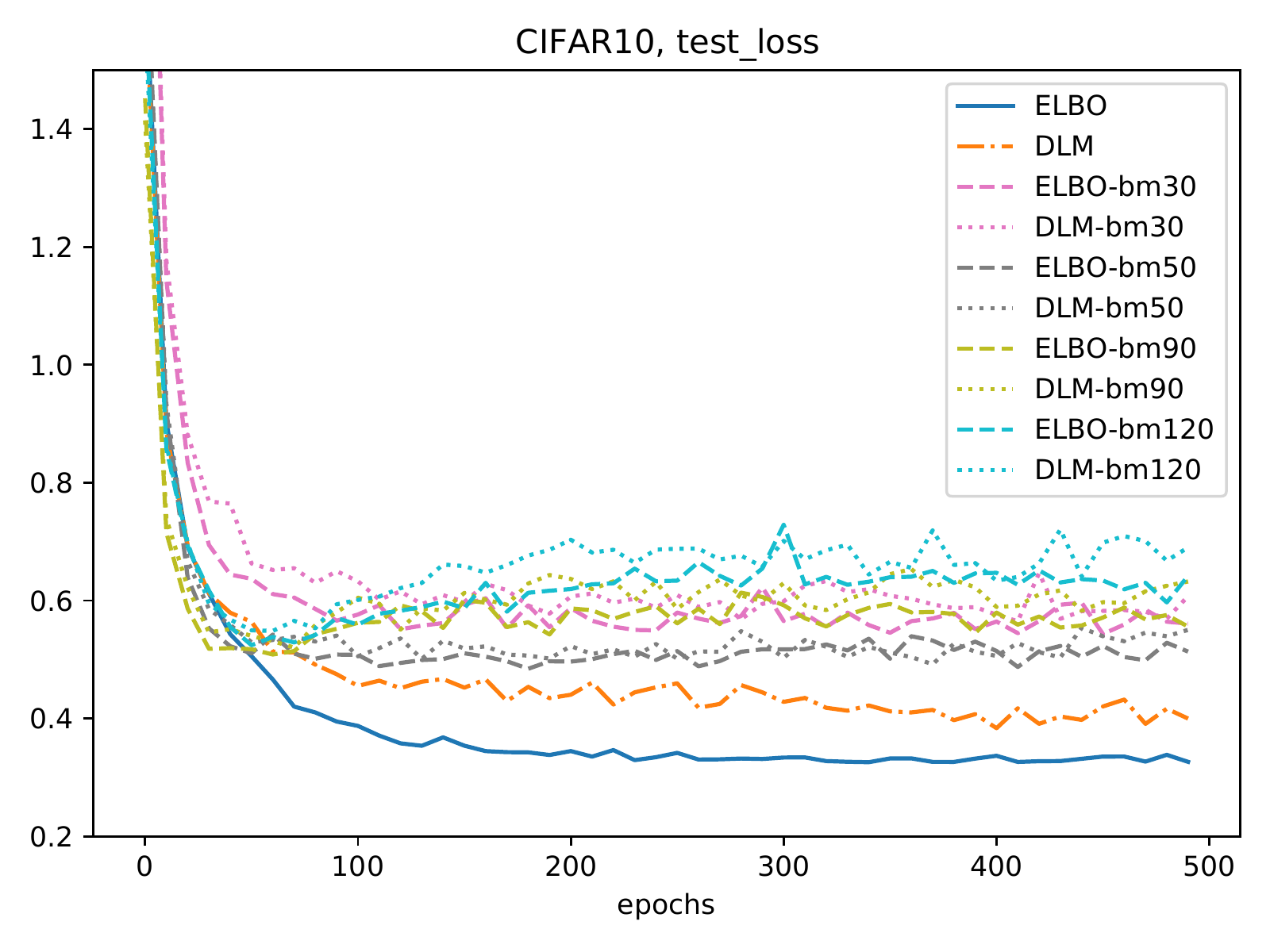}
         \caption{CIFAR10, test loss($\downarrow$)}
    \end{subfigure}
    \begin{subfigure}[t]{0.45\textwidth}
         \centering
         \includegraphics[width=\textwidth]{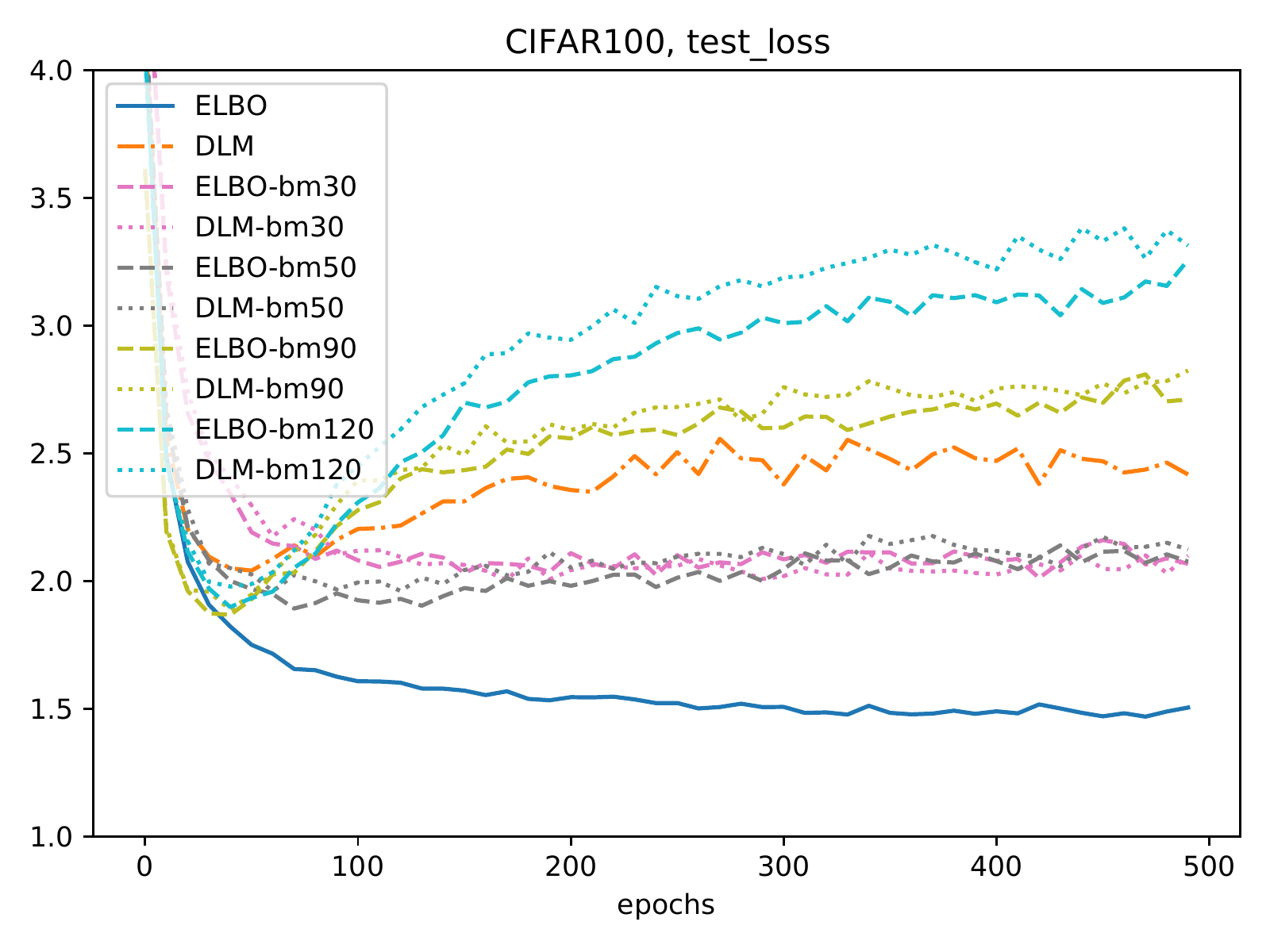}
         \caption{CIFAR100, test loss($\downarrow$)}
    \end{subfigure}
    \caption{Bounded Optimization}
    \label{fig:vibo}
\end{figure}
From Figure \ref{fig:vibo} we can see that the original ELBO still outperforms all other methods, including the original DLM and the solutions with bounded optimization. One observation is that with bounded optimization, DLM is closer to ELBO. 

\subsection{Collapsed Varitional Inference}
\cite{collapsed-elbo} proposed a hierarchical model and performed variational inference on the prior parameters as well. The variational distribution of the prior parameters can be optimized and by marginalizing over prior parameters, a new evidence lower bound is derived. Let $D$ be the total dimension of the weights. The new evidence lower bound replaces the original regularizer, the KL divergence, with a new one:
\begin{align}
    \frac{1}{2 \gamma} [1^\top \sigma^2 + \alpha_{reg} \mu^\top \mu] - \frac{1}{2} 1^\top \log \sigma^2 - \frac{D}{2} \log \alpha_{reg},
    \label{eq:cvi-mean}
\end{align}
if we fix prior variance but learn prior mean (we call this ``mean''), or
\begin{align*}
    (\alpha+\frac{1}{2}) 1^\top \log \left[\beta 1 + \frac{\delta}{2} \mu^2 + \frac{1}{2} \sigma^2 \right] - \frac{1}{2} 1^\top \log \sigma^2,
    \label{eq:cvi-mv}
\end{align*}
if we learn both mean and variance (we call this ``mv'').
We choose $\gamma=0.3, \alpha_{reg}=0.05$ in eq \eqref{eq:cvi-mean}, $\alpha=0.5, \beta=0.01, \delta=0.1$, following the standard settings in \citep{collapsed-elbo}. 

Empirical Bayes \citep{dvi} can be viewed as a special case of collapsed variational inference by restricting the variational distribution of $\sigma^2$ to be a delta distribution and the new regularizer it derives is (we call this ``eb''):
\begin{align}
    \frac{1}{2} \left[D\log \frac{\mu^\top \mu + 1^\top\sigma^2 + 2\beta}{D + 2\alpha + 2} - 1^\top \log |\sigma^2| - D \right] + \frac{1}{2} \frac{D+2\alpha+2}{\mu^\top \mu + 1^\top\sigma^2 + 2\beta} (\mu^\top \mu + 1^\top\sigma^2),
\end{align}
where we set $\alpha=4.4798$ and $\beta=10$ following \cite{dvi}. 

\begin{figure}[H]
    \centering
    \begin{subfigure}[t]{0.45\textwidth}
         \centering
         \includegraphics[width=\textwidth]{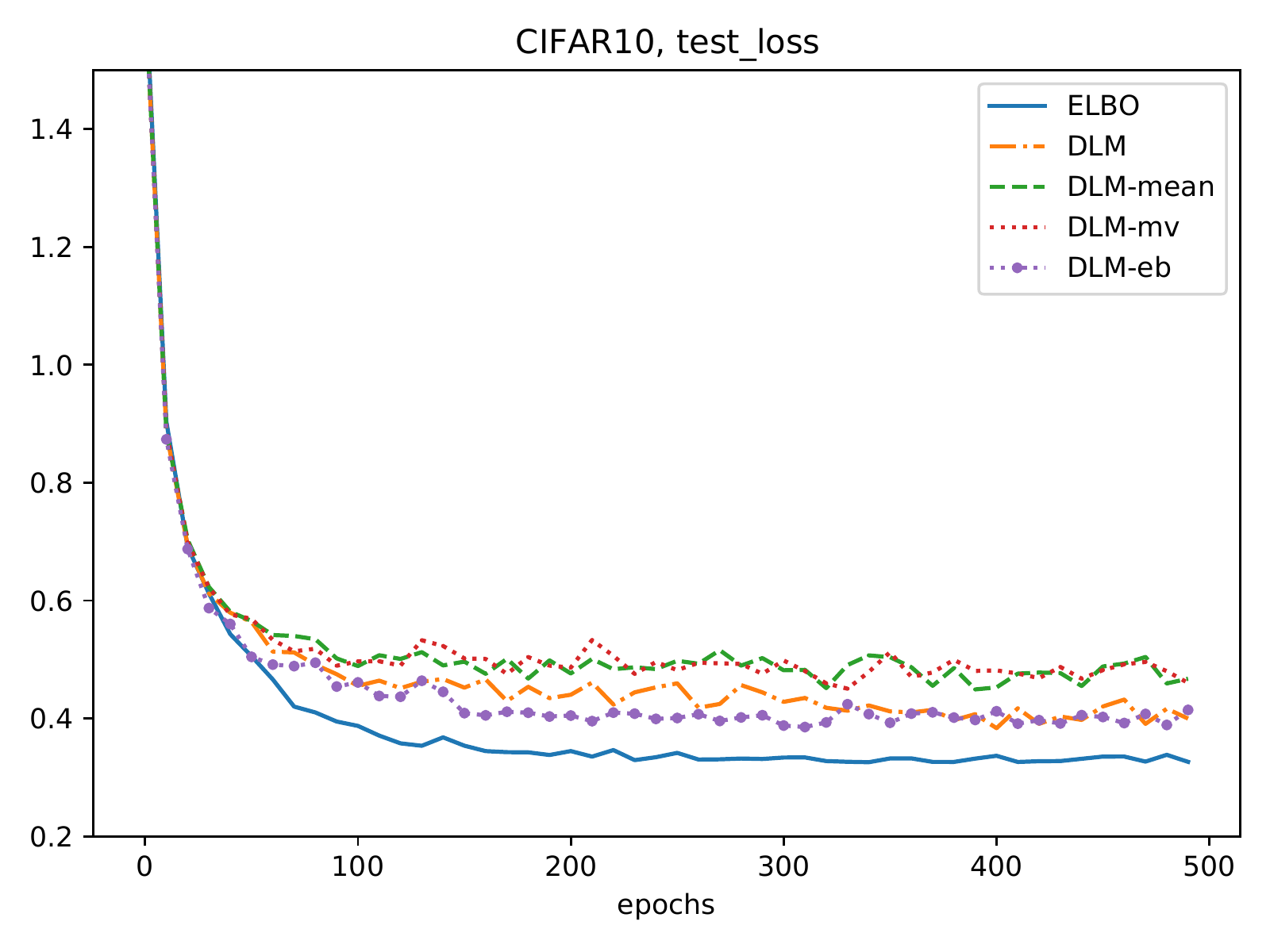}
         \caption{CIFAR10, test loss($\downarrow$)}
    \end{subfigure}
    \begin{subfigure}[t]{0.45\textwidth}
         \centering
         \includegraphics[width=\textwidth]{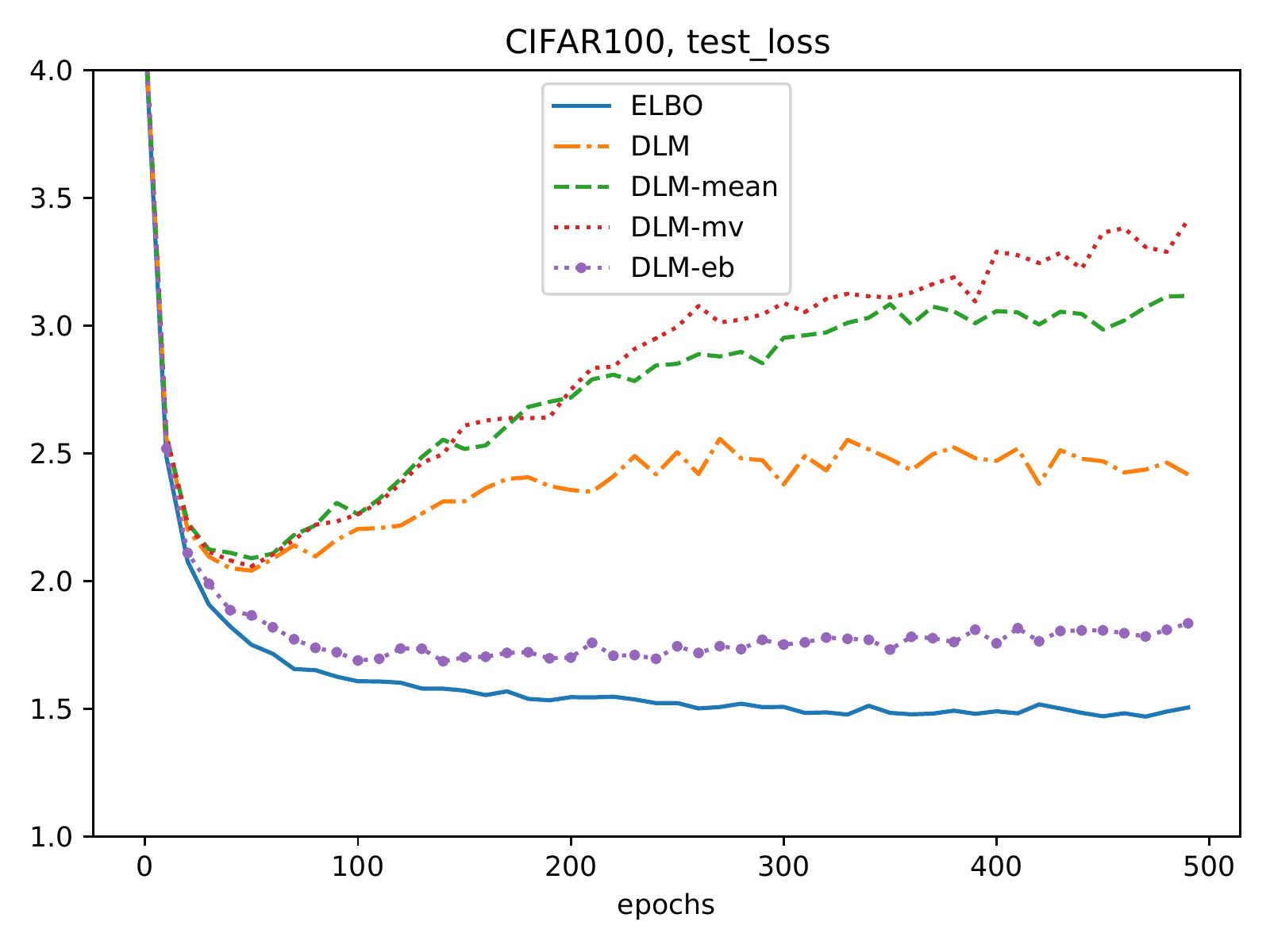}
         \caption{CIFAR100, test loss($\downarrow$)}
    \end{subfigure}
    \caption{Collapsed Variational Inference}
    \label{fig:cvi}
\end{figure}

Figure \ref{fig:cvi} shows the result of applying the new regularizers to DLM. Unlike ELBO, the new regularizers ``mean'' and ``mv'' do not improve the performance of DLM and even make it worse. In contrast, ``eb'' does improve the performance of DLM but this variant of DLM is still worse than ELBO.

\end{document}